\documentclass[a4paper]{article}
\usepackage[margin=1.2in]{geometry}

\usepackage{amsmath,amsfonts,bm}

\def\eqref#1{equation~\ref{#1}}
\def\1{\bm{1}}

\DeclareMathAlphabet{\mathsfit}{\encodingdefault}{\sfdefault}{m}{sl}
\SetMathAlphabet{\mathsfit}{bold}{\encodingdefault}{\sfdefault}{bx}{n}

\def\gD{{\mathcal{D}}}

\def\gX{{\mathcal{X}}}
\def\gY{{\mathcal{Y}}}
\def\gZ{{\mathcal{Z}}}

\newcommand{\E}{\mathbb{E}}

\newcommand{\R}{\mathbb{R}}

\DeclareMathOperator*{\argmax}{arg\,max}
\DeclareMathOperator*{\argmin}{arg\,min}

\usepackage[utf8]{inputenc}
\usepackage{hyperref}
\usepackage[sort&compress,numbers,square]{natbib}
\usepackage{soul} %for crossing out, can be deleted
\usepackage{caption}
\usepackage{subcaption}
\usepackage{pifont}% http://ctan.org/pkg/pifont
\usepackage{amsmath}
\usepackage{amssymb}
\usepackage{amsthm}
\usepackage{mathtools}
\usepackage{booktabs}
\usepackage{comment}
\usepackage{url}
\usepackage{authblk}
\newcommand{\cmark}{\ding{51}}%
\newcommand{\xmark}{\ding{55}}%

\newcommand{\thomas}[1]{{\color{teal} \textbf{Thomas:} #1}}
\newcommand{\tf}{\varphi} % transformation function
\newtheorem{theorem}{Theorem}[section]
\theoremstyle{definition}
\newtheorem{definition}{Definition}[section]

\title{Dissecting Performative Prediction: A Comprehensive Survey}

\author[1]{Thomas Kehrenberg}
\author[1,2]{Javier Sanguino Bautiste}
\author[1,2]{Jose A. Lozano}
\author[1,3,4]{Novi Quadrianto}
\affil[1]{Basque Center for Applied Mathematics, Bilbao, Spain}
\affil[2]{University of the Basque Country UPV/EHU, San-Sebastian, Spain}
\affil[3]{University of Sussex, Brighton, UK}
\affil[4]{Monash University, BSD City, Indonesia}
\affil[ ]{\texttt{\{tkehrenberg,jsanguino,jlozano\}@bcamath.org, n.quadrianto@sussex.ac.uk}}

\date{}

\begin{document}

%%
%% This command processes the author and affiliation and title
%% information and builds the first part of the formatted document.
\maketitle
%%
%% The abstract is a short summary of the work to be presented in the
%% article.
\begin{abstract}
\noindent The field of performative prediction had its beginnings in 2020 with the seminal paper ``Performative Prediction'' by Perdomo et al.,
which established a novel machine learning setup where the deployment of a predictive model causes a distribution shift in the environment, which in turn causes a mismatch between the distribution expected by the predictive model and the real distribution.
This shift is defined by a so-called distribution map.
In the half-decade since, a literature has emerged which has, among other things, introduced new solution concepts to the original setup, extended the setup, offered new theoretical analyses, and examined the intersection of performative prediction and other established fields.
In this survey, we first lay out the performative prediction setting and explain the different optimization targets: performative stability and performative optimality.
We introduce a new way of classifying different performative prediction settings, based on how much information is available about the distribution map.
We survey existing implementations of distribution maps and existing methods to address the problem of performative prediction, while examining different ways to categorize them.
Finally, we point out known and previously unknown connections that can be drawn to other fields, in the hopes of stimulating future research.
\end{abstract}

\section{Introduction}
The word ``performativity'' as we use it in this article stems from the term ``performative utterance''
from the philosophy of language~\cite{austin1975things}.
A performative utterance is a sentence that is primarily not meant to describe reality but to perform an action in the reality.
The mere act of  enunciating a performative utterance changes the word, it does not to describe  the world.
For example, by saying ``I apologize for my actions'', one performs the act of apologizing or ``I promise to do this by tomorrow'', one performs the act of promising.  
% these are not descriptions of anything but they cause you to, for example, be held to a promise.
The concept of performativity  has been widely used in other fields
like finance~\cite{callon1999economists,mackenzie2020economists} (theoretical models of the market change the market)
or journalism (a piece of news can change the opinion of the topic talking about).
In the  context of predictive models, the concept was introduced by \citet{perdomo2020} as \textit{performative prediction}.
Performative prediction refers to a situation
where the act of making a prediction by the model has the potential to change reality.
In particular, we study cases where the prediction affects the part of reality that we were trying to predict --- thereby rendering the prediction inaccurate or producing self-fulfilling prophecies.
This is not as rare an occurrence as one may think.
A prediction of the fastest driving route to some destination can become incorrect when many people heed the prediction such that this route becomes congested.
Automated prediction of credit risk can make customers strategically change financial features to game the predictive model such that it no longer works.
A family can increase the number of books in their household by buying them when they discover that the number of books in their household is an important feature for a model that predicts student's success (because knowledge is valued). This might change their admission in university.   
The last two are an example of Goodhart's Law from economics: ``When a measure becomes a target, it ceases to be a good measure.''
As the output of the classifier becomes important, it becomes a target for manipulation.
However, we do not want to give the impression that performative prediction is merely a re-packaging of Goodhart's Law.
Deliberate manipulation of a measure is not needed in order for the environment to change in response to the deployment of a predictor, as the driving route example shows.

Performative prediction can be understood as a distribution shift in response to the deployment of a predictive model.
In other words, the predictions of the model cause the environment to have a different distribution.
The distribution shift that is induced by a model deployment can be formalized with a \emph{distribution map}, \(\mathcal{D}(\cdot)\),
where \(\mathcal{D}(\theta)\) is the distribution we encounter after deploying a model with model parameters \(\theta\).%
\footnote{Throughout the text we may also use $\theta$ to refer to the \emph{model} with parameters $\theta$.}
The key challenge is that the model parameters \(\theta\) which were optimized on the \emph{initial} data distribution are most likely not optimal any more for the distribution \(\mathcal{D}(\theta)\).
We can retrain the model on the new distribution, but then we will soon have the same problem again and need to retrain again, all the while having a non-optimal model deployed.
Indeed, performative prediction can be seen as a two-step iterative process
where 1) the model is optimized; and 2) there is a distribution shift upon model deployment.
The field of performative prediction is about finding algorithms which produce models that operate well on the distribution that the models themselves are inducing.
Such algorithms may try to anticipate the distribution shift and make the model robust to the shift.
It is therefore important to study both the distribution shift and the optimization procedure. 
In this survey, we will take advantage of this natural division and dissect those two components.
In Section~\ref{sec:distribution-map}, we will revisit the most common mechanisms for the distribution map and propose a novel classification depending on the \emph{levels of access}.
Whereas in Section~\ref{sec:opt-perf-stable} and \ref{sec:opt-perf-opt}, we will discuss the optimization algorithms. 

 Perhaps surprisingly, the distribution map formalism is very broad and  subsumes a lot of other fields of machine learning. Indeed, adversarial attacks or algorithmic recourse --- both being about achieving a different outcome by changing the input as little as possible --- can be seen as performative prediction. Another example is delayed impact of fair machine learning~\cite{liu2018delayed} where fairness interventions have long-term effects on the population which change the predictive landscape. In Section~\ref{sec:related-research-areas}, we will explore how they fall under the umbrella of performative prediction and establish connections with the hope that we inspire future research to find more connections to these more established fields.

\subsection{Three guiding examples}%
\label{ssec:guiding-examples}
The basic setup of a classifier influencing the data distribution is very general.
To guide the reader through the survey, we introduce three examples that we will be referring to repeatedly.

\subsubsection{Bank loan decisions}
The first example is from \citet{perdomo2020}:
We consider a bank that wants to make automated lending decisions.
The bank trains a classifier on past data of the history of defaulting, \(y\in\{0,1\}\), based on available financial features, \(x\in \mathbb{R}^d\) of the clients. The classifier predicts the default risk $\hat{y}\in[0,1]$. 
This model is then applied to prospective customers
and if the predicted default risk is below some threshold, the loan is granted.
This usage of the model can have two performative effects:
The first is that if the model predicts high default risk for certain customers,
those customers may get higher interest rates, which in turn increases the probability of default,
making the model incorrectly underestimate the default risk.
This problem could potentially be solved by making default risk predictions conditioned on the interest rate.
However, there is another, harder-to-fix, problem:
the customers themselves may react to the existence of the model by strategically changing their financial features ---
for example, changing their payment patterns.
This invalidates the correlations that the model originally learned from the historical data.

\subsubsection{Graduation rate intervention}\label{sssec:graduation-rate}%
    This example is from \citet{kim2023making}.
    Some school districts in the U.S. have a system for predicting the likelihood of graduation for high-school students,
    in the context of ensuring higher graduation rates.
    If the probability is low for a particular student,
    interventions are performed by the schools in order to improve graduation chances.
    Thus, we have a predictive model, which predicts whether someone will graduate --- call this prediction \(\hat y\),
    and this predictive model influences the true graduation outcome, \(y\),
    via the fact that resources will be spent on improving the chances of high-risk students.
    In this case, the institution \emph{intends} the performative effect --- it wants high-risk students to have a better chance ---
    but the performative effect nevertheless provides a unique challenge during training and evaluation of the predictive model.
    This example is a case of \emph{outcome performativity} --- a special case of performative prediction
    that we will examine in more detail below.

\noindent

% (what is the purpose of the following example?)

\subsubsection{Car navigation system}
Consider a navigation system which predicts the fastest route based on the current traffic.\footnote{This example is inspired by \citet{perdomo2020}.}
For each possible route, it assigns a prediction for how long the drive will take.
Imagine the scenario of leaving a concert or a massive event:
if many people see the output of this system and decide to take the route that is predicted to be the fastest,
the route may become so congested that it is not the fastest route any more.
This could, for example, happen if the main road is partially blocked due to an accident and there is a smaller cross-country road as an alternative that naïvely seems faster.
Thus, the model's prediction has changed the environment --- via the decisions of the drivers --- and can no longer make accurate predictions about the environment.
\section{The basic setup}\label{sec:2}

\subsection{All the world's a stage: characterizing performativity}% (fold)
\label{ssec:performativity}

At its heart, \emph{performative prediction} (PP) is about the fact that once you deploy a classifier,
the data distribution you encounter afterwards can change because the environment reacts to your classifier.
Let \(\gZ\) be the space of data you encounter.
A classifier with parameters \(\theta\in\Theta\) is trained to minimize a loss on \(z\in\gZ\): \(\ell(z; \theta)\), for example a binary cross-entropy loss.
Let \(\gD_\mathrm{init}\) be the initial distribution of data,
then, following standard empirical risk minimization, the optimal classifier on that distribution, \(\theta^*_\mathrm{init}\),
is the one that solves (not necessarily uniquely) the following optimization problem:
\begin{equation}
    \theta^*_\mathrm{init} = \argmin_{\theta\in\Theta}\; \mathbb{E}_{z\sim \mathcal{D}_\mathrm{init}}[\ell(z; \theta)]
    = \argmin_{\theta\in\Theta}\; \int_{\mathcal{Z}}\ell(z; \theta)\,p_ \mathrm{init}(z)\,\mathrm{d}z
\end{equation}
where \(p_ \mathrm{init}(z)\) is the probability density corresponding to the distribution \(\mathcal{D}_ \mathrm{init}\).
% where \(\mathbb{E}_{z\sim \gD_\mathrm{init}}[\cdot]\) represents the expectation value over the distribution \(\mathrm{init}\).
Note that \(\gZ\) is general enough to represent a variety of settings. For example, for supervised learning it would typically manifest as the Cartesian product of a feature space \(\gX\) and a set of labels \(\gY\)
(i.e.\ \( \gZ \subseteq \gX \times \gY\)).

So far, this is an unremarkable ML setup,
but the interesting aspect of the PP setup is  that the data distribution changes ---
conditional on the deployed classifier, \(\theta\).
Concretely, there is a mapping, \(\gD(\cdot)\), called the \emph{distribution map},
from the parameter space to the space of distributions over \(\gZ\), which we call \(\Delta(\gZ)\):
\begin{equation}
\label{eq:distribution-map}
    \gD: \Theta \to \Delta(\gZ)~.
\end{equation}
If \(\theta^*_ \mathrm{init}\) refers to the model trained on the initial distribution, \(\gD_\mathrm{init}\),
then the distribution changes to be \(\gD(\theta^*_ \mathrm{init})\) once the model is deployed.
The expected loss of the model with parameters \(\theta^*_ \mathrm{init}\) may now be much higher on the induced distribution \(\gD(\theta^*_ \mathrm{init})\),
than on the data distribution that it was trained on.

% CANDIDATE FOR DELETION
This is one of the core challenges of performative predictions:
finding a model that is not necessarily the best on the initial distribution,
but which is best on the distribution that it itself induces.
This will be made more formal in Section~\ref{ssec:performative_optimality_and_stability}.

The setup can be seen as a special case of a Markov decision process (MDP) as studied in reinforcement learning,
where the actions are the possible models \(\theta\) that can be deployed,
the (deterministic) transition function is the distribution map, \(\mathcal{D}(\theta)\),
and the reward is the risk (as defined below in Section~\ref{ssec:performative_optimality_and_stability}).
However, note that there is no statefulness or memory in performative prediction:
the environment always reacts to \(\theta\) as if it were the first model deployed.
This is in contrast to the stateful PP setting introduced later on.

    Performative prediction is thus closer to a (continuum-armed) bandit problem~\cite{agrawal1995continuum,kleinberg2004nearly,auer2007improved,kleinberg2008multi,podimata2021adaptive}.
    However, as pointed out by \citet{jagadeesan2022regret}, in contrast to a bandit problem, we can observe more about the effects of our actions (that is, the deployment of our model \(\theta\)) than just the reward (that is, the risk):
    we can observe the induced distribution \(\mathcal{D}(\theta)\).
    The connection to bandit problems will be explored in detail in Section~\ref{sssec:bandits-assuming-sensitivity-and-smoothness}.
    A variations of performative prediction that is \emph{stateful} (and thus closer to an MDP) will be discussed in Section~\ref{ssec:stateful-performative-prediction}.
    % This statelessness of performative prediction invites different solution concepts than those used on MDPs.

\subsection{Performative optimality and stability}%
\label{ssec:performative_optimality_and_stability}
For a given loss function, \(\ell(z, \theta)\),
% and a given distribution map, \(\gD\),
we define the %(empirical) 
\emph{risk}, the expected loss of the model specified by \(\theta\) on the data distribution \(Q\), as:
\begin{equation}
  % \gL(\theta, \theta') := \E_{z\sim \gD(\theta')}[\ell(z; \theta)]~.
  \mathrm{Risk}(\theta, Q) = \E_{z\sim Q}\big[\ell(z; \theta)\big]~.
  \label{eq:risk}
\end{equation}
%
% The risk is parametrized by two model parameter value vectors: the first,
% \(\theta\), defines the model for which the loss is computed, and the second,
% \(\theta'\), is passed to the distribution map and determines the data
% distribution.
Recall that the distribution map, \(\gD(\cdot)\), maps the parameters of a model, \(\theta\), to the distribution induced by model \(\theta\).
Thus, with \(Q:=\gD(\theta')\), \(\mathrm{Risk}(\theta, \gD(\theta'))\) describes the risk of the model defined by \(\theta\) on the distribution induced by \(\theta'\).
With these definitions, there are two natural optimization targets one can define:
\emph{performative stability} and \emph{performative optimality}.

The model parameters \(\theta_ \mathit{PS}\) are \emph{performatively stable} if they
minimize the risk on the distribution induced by \(\theta_ \mathit{PS}\) itself:
\begin{equation}
  \theta_\mathit{PS}=\argmin_{\theta\in\Theta} \;\mathrm{Risk}\big(\theta, \gD(\theta_ \mathit{PS})\big)~,
  \label{eq:perform-stable}
\end{equation}
meaning that if we find ourselves confronted with the distribution \(\gD(\theta_ \mathit{PS})\), then the optimal model that we will find for this distribution is $\theta_\mathit{PS}$ itself.
In other words, \(\theta_ \mathit{PS}\) is a fixed point for the function
\(g(\theta)=\arg\min_{\theta'\in\Theta} \mathrm{Risk}(\theta', \gD(\theta))\).
For a well-behaved (by which we mean sufficiently (strongly) convex and Lipschitz) \(\ell\) and \(\gD(\cdot)\),
\(\theta_ \mathit{PS}\) is guaranteed to be found by repeated application of $g(\cdot)$.
(See Section~\ref{sec:opt-perf-stable} for more details.)

On the other hand, the model parameters \(\theta_ \mathit{PO}\) are said to be \emph{performatively optimal}
if they result in the overall lowest risk when evaluated on the distribution the model itself induces:
\begin{equation}
  \theta_\mathit{PO}=\argmin_{\theta\in\Theta} \;\mathrm{Risk}\big(\theta, \gD(\theta)\big)~.
  \label{eq:perform-optim}
\end{equation}
This definition immediately implies
\begin{equation}
   \mathrm{Risk}\big(\theta_\mathit{PO}, \gD(\theta_\mathit{PO})\big)\leq 
   \mathrm{Risk}\big(\theta_\mathit{PS}, \gD(\theta_\mathit{PS})\big)~.
  \label{eq:risk-opt-vs-stbl}
\end{equation}
Crucially, $\theta_\mathit{PO}$ is \emph{not} a fixed point in general,
but rather, the global minimum for the \emph{performative risk}, $\mathit{PR}$,
which is defined as follows:
\begin{equation}
  \mathit{PR}(\theta) = \mathrm{Risk}\big(\theta, \gD(\theta)\big)= \E_{z\sim \gD(\theta)}[\ell(z; \theta)]~.
  \label{eq:perform-risk}
\end{equation}
The performative risk is characterized by the fact that the same model parameters are passed to the loss and the distribution map.

In the general case, optimality and stability are unrelated --- neither implies the other.
However, as we will see in later sections, under certain assumptions,
the stable point is a good approximation for the optimal point.

With all the important concepts now introduced, we can draw a comparison between performative prediction and \emph{Stackelberg games}, a strongly related concept from economics.
Appendix~\ref{appendix:stackelberg-games} gives a brief overview of the concept of Stackelberg games and explains how it maps to performative prediction.
Stackelberg games have been thoroughly studied in economics but the economic literature focuses on analytic solutions to the problem, which are not applicable in ML\@.

\begin{figure}
    \centering
    \includegraphics[width=\textwidth]{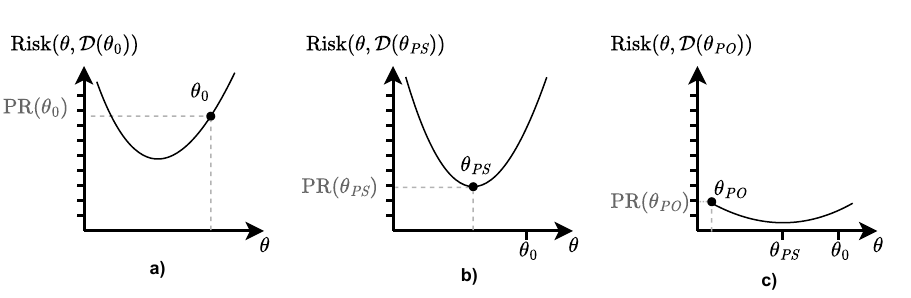}
    \caption{%
        Let $\ell(z;\theta) = z\cdot (\theta^2+1)$ and $\mathcal{D}(\theta) = \mathcal{N}(\sqrt{a_1\theta + a_0}; \sigma)$ with $a_0,a_1\in\mathbb{R}$. The figure shows three examples of
        picking a distribution and plotting the corresponding risk: \(\gD(\theta_0)\), \(\gD(\theta_\mathit{PS})\), and \(\gD(\theta_\mathit{PO})\).
        In the first plot, \textbf{a)}, we fix the distribution to $\mathcal{D}(\theta_0)$ where $\theta_0$ is an arbitrary starting point.
        To show the performative risk, we mark $\mathrm{Risk}(\theta_0, \mathcal{D}(\theta_0))$.
        We see that $\theta_0$ not stable, as it is not the minimum of the risk for the distribution that $\theta_0$ induces.
        The second plot, \textbf{b)}, shows the risk for a stable point $\theta_\mathit{PS}$.
        We can certify that it is stable, because for the given distribution $\mathcal{D}(\theta_\mathit{PS})$, it achieves the lowest risk.
        In general, there can be more than one stable point.
        Finally, plot \textbf{c)} shows the optimal point, $\theta_\mathit{PO}$.
        We cannot tell from the plot that this is the optimal point (see Fig.~\ref{fig:opt} for a different plot where we are able to tell).
        As the plot shows, the optimal point does not necessarily produce the lowest risk for its own distribution $\mathcal{D}(\theta_\mathit{PO})$,
        but it produces the lowest performative risk of all $\theta\in\Theta$. In other words, the optimal point need not to be stable.
    }%
    \label{fig:opt-vs-stable}
    % Description of the figure for visually-impaired readers:
\end{figure}

\begin{figure}
    \centering
    \includegraphics[width=\linewidth]{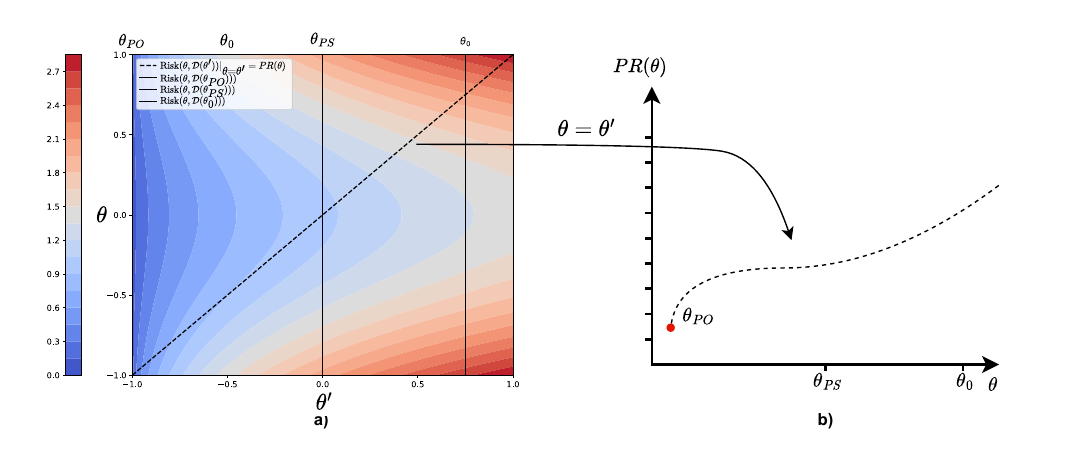}
    \caption{%
        Figure \textbf{a)} shows a contour plot (elevation view) of the 3D surface corresponding to stacking all possible risk curves, $Risk(\theta,\mathcal{D}(\theta')), \forall\theta,\theta'\in\Theta$, for the example used in Fig.~\ref{fig:opt-vs-stable}. Note that Fig.~\ref{fig:opt-vs-stable} shows three instances of such curves; the corresponding section of the surface of these instances is marked in the contour plot. The optimal point is the minimum of the section where $\theta=\theta'$ (marked with a dotted line) i.e. $\theta_{PO}=\argmin_{\theta \in \Theta} PR(\theta) =\argmin_{\theta \in \Theta} \mathbb{E}_{z \sim \mathcal{D}(\theta)}[\ell(\theta;z)]$. The corresponding performative risk with its minimum is shown in \textbf{b)}. 
        The difference between plot \textbf{c)} in Fig.~\ref{fig:opt-vs-stable} and plot \textbf{b)} in this figure is that the former shows the risk for the fixed distribution $\mathcal{D}(\theta_\mathit{PO})$, whereas the latter shows the performative risk, $\mathit{PR}$.
    }%
    \label{fig:opt}
    % Description of the figure for visually-impaired readers:
\end{figure}

Finding the optimal point is usually much harder,
because it requires exploring the entire distribution map, \(\gD(\cdot)\),
even when the distribution map is very smooth.
Conversely, an iterative procedure can usually find the stable point quickly,
if we again assume sufficient smoothness in the distribution map.
Iterative procedures are only guaranteed to find the optimal point
if the performative risk as a whole is convex~\cite{miller2021outside},
which is a much stronger assumption.

Finding the optimal point is typically more desirable because it lets us achieve the lowest possible risk (see \eqref{eq:risk-opt-vs-stbl}).
Additionally, optimality is difficult to \emph{certify}.
To check whether a model parameter vector \(\theta\) corresponds to a \emph{stable} point,
we merely have to check that it is the result of risk minimization on its own induced distribution.
Having a stable point is thus the property of a single model.
For the optimal point, on the other hand,
we have to consider all possible model parameters \(\theta\) and verify that the optimal point performs better on its own induced distribution
than the other model parameters do on their own induced distributions.
This difference between optimal and stable point is illustrated in Fig.~\ref{fig:opt-vs-stable} where the stable point is easy to identify while the optimal point is not.
Fig.~\ref{fig:opt} shows conceptually how to find the optimal point: by varying $\theta$ in the performative risk, which corresponds to the ``diagonal'' in the $\Theta\times\Theta$ space that spans all possible values of $\mathrm{Risk}(\theta,\mathcal{D}(\theta'))$ (see the figure caption for details).
% The optimal point can be seen as the best-performing point in the closed-loop interaction between the model and the distribution map, as Fig.~\ref{fig:opt} shows.

\subsection{Performative Prediction is an iterative process}%
\label{sec:two-phases}

\begin{figure}
    \centering
    \includegraphics[width=\textwidth]{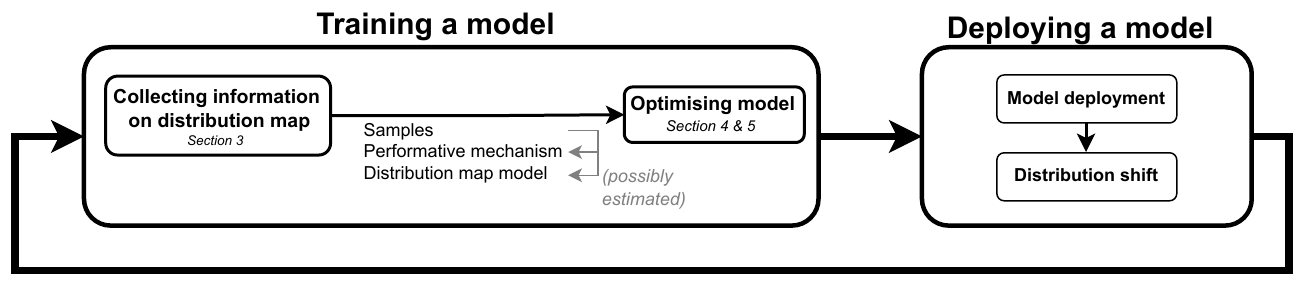}
    \caption{%
            Training a model in the PP framework involves first collecting information on the distribution map.
            In the simplest case, we just collect samples and train the model on the samples.
            However, some methods for addressing performative prediction need more than samples:
            they need a mathematical model of the performative mechanism or of the whole distribution map
            (see the beginning of Section~\ref{sec:opt-perf-opt} for an explanation).
            To go from samples to a mathematical model of the distribution map,
            an estimation mechanism (discussed in Section~\ref{ssec:distribution-map-model-fitting}) can be used,
            which, however, incurs an estimation error.
            With a mathematical model of the distribution map,
      it is possible to apply optimization algorithms to \eqref{eq:risk} when $Q$ depends on $\theta$.
      We discuss algorithms to find the stable point and optimal point in Sections~\ref{sec:opt-perf-stable} and \ref{sec:opt-perf-opt} respectively.
      Once the model is trained, it is deployed, causing a distribution shift and the necessity of retraining the model. 
    }%
    \label{fig:two-step}
    % Description of the figure for visually-impaired readers:
\end{figure}

As discussed, performative prediction can be seen as a two-step iterative problem where (i) first the institution trains a model; and (ii) then, upon model deployment, the data distribution shifts. 

Typically, ML only focuses on the optimization problem that training a model represents.
However, in this case, we also have to attend to the distribution map.
% The distribution shift is a direct consequence of deploying a certain model. Nevertheless, this does not mean that the attention should be completely off the distribution map.
In order to solve the optimization problem, some information about the distribution map has to be obtained as it captures the information of the underlying data distribution in the machine learning problem. Therefore, to solve any performative prediction problem, one should (i) get information about the distribution map (treated in Section~\ref{sec:distribution-map}) and (ii) solve the optimization problem (treated in Sections~\ref{sec:opt-perf-stable} and \ref{sec:opt-perf-opt}), as Fig.~\ref{fig:two-step} illustrates. 

We use this inherent division of components in performative prediction to categorize previous work and to structure this survey. Section~\ref{sec:distribution-map} reviews and categorizes the \textit{distribution maps} previously used and Sections~\ref{sec:opt-perf-stable} and \ref{sec:opt-perf-opt} review \textit{optimization algorithms} to reach the stable and optimal point, respectively. 

\subsection{Operational aspects of performative prediction}%
\label{ssec:practical-concerns}
In this section, we will look at  aspects of performative prediction which the mathematical definition above leaves unspecified.

\subsubsection{Where do the training samples come from?}%
\label{sssec:distribution-map-known}
Which strategy to choose for addressing performative prediction depends on how well the distribution map, \(\gD(\cdot)\),
is known to the model deployer.
If it is, for example, known in sufficient detail to write an efficient simulation of it,
then the model deployer can simulate the response of the environment over as many iterations as desired,
and thus can train a (close to) optimal model without interacting with the real world.
In this situation, a simple iterative-retraining strategy would work well, for example.
%
% This setup corresponds almost exactly to that of \emph{Stackelberg games},
% which we describe in Section~\ref{ssec:stackelberg-games}.
%

On the other hand,
if no good simulation of the distribution map exists,
significant time will pass where a model is deployed and the new distribution is observed,
until --- based on the feedback from the environment --- enough information from the distribution map is obtained to optimize a new model.
In such circumstances a simple retraining strategy might take too long and might be too expensive.

    We can distinguish between the two basic cases:
(i) expensive sample access after real deployment;
% we need to limit the amount of samples we use to build our model.}
and (ii) cheap sample access through (black-box) simulation.
% we can draw an unlimited number of (simulated) samples from the distribution map.}
% but do not have white-box access.

However, the line between cases (i) and (ii) can be difficult to draw,
because in past work, it has not been clear where samples are coming from:
the real world (case (i)) or a simulation (case (ii)).
This is not just an academic question, because methods which need a large number of samples --- like repeated risk minimization (Section~\ref{ssec:repeated-risk-minimisation}) ---
seem only viable with cheap sample access, i.e., case (ii).

We will pick up this thread again later in Section~\ref{sec:distribution-map} when we discuss the ways to simulate a distribution map in benchmarks and experiments.

\subsubsection{How quickly do users react to the newly deployed model?}%
\label{sssec:stateful-distribution-maps}
The formalism described above does not specify how long it takes users to react to a new model \(\theta\).
The distribution map formalism simply states that at some point after the deployment of the model,
the distribution becomes \(\mathcal{D}(\theta)\) and that the distribution does not change any more after that (until the next model deployment).
In other words, the canonical performative prediction setup assumes an \emph{equilibrium state} which is reached after some unspecified amount of time.
The time until new equilibrium can be short, as in the car navigation system:
Traffic will likely come to a new equilibrium within hours after GPS systems have started recommending different routes.
In other examples, the deployer may have to wait years until the environment has settled on a new distribution.
Consider the loan decision example, not all applicants will understand immediately how the model works,
and so there will be a time period of adaptation where more and more applicants learn how to best react to the deployed model.

There are two possible problems  for using performative prediction methods in the gradual adaptation scenario.
The first problem only appears if we have to rely on real samples, i.e., case (i) as described above in Section~\ref{sssec:distribution-map-known}.
In this case, the method might take an impractical amount of time to find a good model.
We know of no solution to this problem.

The second problem is more subtle.
If we assume access to \emph{cheap samples} (case (ii) in Section~\ref{sssec:distribution-map-known}), then waiting for equilibrium is not needed:
we have a mathematical model of the distribution map, which, by assumption, returns the equilibrium state.
As such, we can immediately deploy the model that is optimal for the equilibrium state.
However, when this model is deployed, the equilibrium state has not been reached yet,
so the model is not optimal during the time it takes for the equilibrium state to be reached.
For this reason, \citet{brown2022performative} have proposed a method for finding intermediate models: models that are optimal for the sections of time before the equilibrium has been reached.
This slow adaptation scenario cannot be modelled with a distribution map that only depends on the last deployed model, \(\theta\), because the distribution map also needs to know how far the adaptation has progressed.
The authors assume fixed time steps and model this as a distribution map which depends on the distribution of the last time step.
They call this new distribution map a \emph{transition map}: \(\mathrm{Tr}:\Theta\times\Delta(\mathcal{Z})\to \Delta(\mathcal{Z})\)
and use it to explore gradual adaptation.
Importantly, with this modification of the setup, the current data distribution does not only depend on the last deployed model, \(\theta\),
but also on the previous data distribution, and thus depends on the \emph{trajectory} of data distributions.
The transition map therefore describes a setup which has been called \emph{stateful performative prediction}.
As this variant has its own dedicated literature, we discuss it in more detail in Section~\ref{ssec:stateful-performative-prediction}.

\subsubsection{Does the distribution map also change the true label?}%
\label{sssec:reuse-labels}
The definition of the performative prediction setup is  given in terms of a single data space \(\gZ\).
However, in many concrete examples, the data will be the Cartesian product of the user features, \(\gX \subset \R^d\),
and labels, \(\gY = \{0, 1, \dots\}\).
In the case of the loan example,
the features are financial data about the applicant and the label indicates whether the applicant is a default risk.
We thus sample two variables: \((x,y)\sim \gD(\theta)\).
%, and use the binary cross-entropy loss:
%
%\begin{equation}
%
%  \ell((x, y); \theta) = y\cdot \log\big(h_\theta(x)\big) +
%  (1-y)\cdot\log\big(1-h_ \theta(x)\big)
  %
%  \label{eq:cross-entropy}
%
%\end{equation}
%
%where $h_ \theta$ is the model with parameter values \(\theta\).

% \thomas{the following two paragraphs need to be rewritten (maybe mention algorithmic recourse)}
Now, where do the labels come from?
We can again distinguish between cases where a (possibly black-box) \emph{simulation} of the distribution map is known (case (ii) according to Section~\ref{sssec:distribution-map-known}) and cases where all samples are obtained from deployment in the real world (case (i)).

If we have a known simulation of the distribution map, then that simulation will also give us information about how the labels behave after the deployment of a classifier;
the labels may change or they may stay the same.
On the other hand, if  we  collect samples from the real world,
the sourcing of the labels becomes a challenge.
If we again consider the loan example,
we can observe the new features of the applicants after we have deployed our initial model,
but we cannot easily observe the labels --- e.g., did any of the feature changes actually change the risk of default?.
Indeed, if we had a reliable way of determining the labels automatically,
then there would be no need for the classifier.
So, all that is left is expensive human labelling of the new data,
which may require waiting several years to actually observe the rate of defaults.

    A notable special case of performative prediction is \emph{outcome performativity}~\cite{kim2023making} where \emph{only} the true label, \(y\), changes and the features, \(x\), remain unchanged.
    Formally, this means that the marginal distribution over \(x\) does not change after the distribution shift.
    However, the conditional distribution \(P(y|x)\) does change.
    An example for this is our \emph{graduation rate intervention} example (\ref{sssec:graduation-rate}):
    after a student was predicted to be at high risk of not graduating,
    we can imagine that there is no change in the original features (which might have been their living situation and their past academic performance, which cannot be affected anymore),
    but their true chance of graduation, \(y\), may nevertheless have changed because the school will spend more resources on them going forward.
    % In a case like this, it can be feasible to learn the performative mechanism (i.e., the distribution map) from observational data, such that we can predict \(y\), conditioned on a previous prediction \(\hat y\) (see section ???).
    We will define \emph{outcome performativity} formally in Section~\ref{sssec:outcome-perf-def}.

\subsubsection{How well can users know the institution's model?}%
\label{sssec:dependence-on-the-model-weights}
As formulated above, the distribution map, \(\mathcal{D}\), depends on the exact parameter values of the model, \(\theta\).
If we translate this statement into a real-world setting, it means that users can base their decisions on the exact model parameters of the institution's model.
Defining \(\mathcal{D}\) like this makes some of the theoretical analysis easier,
but it is for most situations not a wholly realistic assumption.
In most real-world scenarios,
the model itself would be kept private by the institution and users would only observe predictions from the model.
(Note that we are now taking the perspective of the \emph{user} and observe samples from the \emph{model};
whereas in Section~\ref{sssec:distribution-map-known} we took the perspective of the institution and were observing samples from the \emph{distribution map}.)

As \citet{izzo2022learn} put it (in section 3.1 of their paper):
``Unless the population being modelled consists mostly of data scientists, it is unlikely that the constituent individuals will have a reaction based on the particular parameters of the model.''
Rather, it is more realistic that users base their reaction on low-dimensional statistics of the model behaviour~\cite{izzo2022learn}.

Consider, for example, the car navigation example:
the new traffic distribution does not depend on the exact parameter values of the car navigation's prediction model,
but rather only depend on the output of the prediction model (by telling each individual driver where to go).

Several works have explored variations on the basic \emph{performative prediction} setup,
where the exact model parameter values are not assumed to be known to the users.
In~\citet{mofakhami2023performative},
the distribution map is assumed to be a function that depends on the \emph{functional behaviour} of the model parameterized by \(\theta\).
That is, if there are two sets of parameters, \(\theta\) and \(\theta'\),
such that the corresponding model behaviour is indistinguishable --- the same inputs lead to the same outputs in both -- this implies that \(\mathcal{D}(\theta)=\mathcal{D}(\theta')\).
In other words, if the models behave the same, they induce the same response in the environment.
This is not necessarily the case in the usual formulation of performative prediction.
% That is, if \(f_\theta\) describes the entire input-output behavior of the model given by \(\theta\),
% then we can write \(\mathcal{D}(f_\theta)\) for the distribution this model induces.
% %
% We can conceptualize \(f_\theta\) as the set of all input-output pairs: \(\{(x,y)|f_\theta(x)=y\}\).
% %
% This is not a strong assumption, but it leads to a different analysis
% in terms of smoothness conditions for convergence guarantees (see Section~\ref{sssec:convergence-model-output-dist-map}).

%
    \emph{Outcome performativity}~\cite{mendler2022anticipating,kim2023making} --- a special case of performative prediction --- similarly assumes that only the \emph{output} of the institution's predictive model is relevant for the distribution shift.
    However, outcome performativity goes further than that.
    Outcome performativity assumes a kind of \emph{locality}:
    whether or not a tuple \((x, y)\) from the training set is affected by performativity
    depends only on \(\hat y = f(x)\), where $f$ is the institution's predictive model.
    Importantly, what happens to \((x, y)\) under the distribution shift does \emph{not} depend on the model's output on any \emph{other} input \(x'\),
    the way it might in \citet{mofakhami2023performative}'s setup.
    If we go back to our \emph{graduation rate intervention} example (\ref{sssec:graduation-rate}),
    then we can see that this locality holds there (in the idealized case):
    whether the school directs additional resources to a student, depends only on the risk score of that particular student and no-one else's score.
    (Of course, in the real world, a school's resources are finite and so if one student receives help, another student might receive less help --- thus breaking locality --- but this effect is ignored in \citet{kim2023making}.)
    Note that this is not the only assumption that defines \emph{outcome performativity}; see Section~\ref{sssec:outcome-perf-def} for a formal definition.

Another work worth mentioning here is~\citet{ghalme2021strategic},
which considers a scenario in which even less information about the model is known to the users.
In one of the scenarios considered there,
the users have to approximate the true model based on finite samples that they can observe.

Allowing the distribution map to depend on the exact value of \(\theta\) is a strictly more general setup, so this will be the assumption throughout most of this survey.

\subsection{Defining objectives in the stateful setting}%
\label{ssec:stateful-performative-prediction}
    We mentioned above in Section~\ref{sssec:stateful-distribution-maps}
    that in order to model gradual adaptation to the deployment of the model,
    \citet{brown2022performative} introduced a variant of performative prediction which is \emph{stateful}.
    The idea is that the distribution shift does not only depend on the deployed model, but also on the previous data distribution.
    This is modelled with the transition map:
    \begin{align}
        \mathrm{Tr}:\Theta\times\Delta(\mathcal{Z})\to \Delta(\mathcal{Z})~.
    \end{align}
    The intuition here is that, for example, in the bank loan example,
    applicants try to modify their feature values to receive a favourable outcome,
    but in each time step they can only move the values a certain amount from the value they had in the previous time step.

    We mentioned previously how performative prediction can be understood as a Markov decision process (MDP), but where the MDP is \emph{stateless}.
    \citet{brown2022performative}'s variant of performative prediction is not stateless and fits the template of an MDP better:
    % This is reminiscent of a Markov decision process as studied in reinforcement learning,
    % where there are states, actions, transition probabilities, and rewards.
    %
    The state is the current distribution over \(\mathcal{Z}\),
    the actions are the possible models to deploy, \(\theta\), which cause a distribution shift,
    the transition probability is deterministic and given by \(\mathrm{Tr}\),
    and the reward is the performative risk, $\mathit{PR}(\theta)$, given in \eqref{eq:perform-risk}.
    However, even stateful PP can be more efficiently addressed by specifically-tailored algorithms than by employing reinforcement learning.
    The reason is the same as for why (non-stateful) performative prediction is not just a bandit problem:
    at each step we can observe samples from the current data distribution, which allows us to gain relevant information about the best models to deploy.
    %as we will see in Section~\ref{ssec:stateful-performative-prediction}.

%
    It is possible to define stability and optimality objectives in this setting as well.
    As a prerequisite, we define what it means to be a \emph{fixed point distribution}:
    A distribution \(Q\in \Delta(\mathcal{Z})\) is a fixed point for \(\theta\) if
    \begin{align}
        \mathrm{Tr}(\theta, Q) = Q~.
    \end{align}
    A distribution-classifier pair, \(Q_ \mathit{SPS}, \theta_\mathit{SPS}\), is a \emph{stable pair} if:

\begin{enumerate}
    \item \(Q_\mathit{SPS}\) is a fixed point distribution for \(\theta_\mathit{SPS}\), and
    \item \(\theta_\mathit{SPS}= \argmin_{\theta\in\Theta}\mathrm{Risk}(\theta,Q_\mathit{SPS})\), i.e., \(\theta_\mathit{SPS}\) minimizes the loss on \(Q_\mathit{SPS}\).
\end{enumerate}
    The first condition implies that the users are no longer adapting
    and the second condition is roughly the same as the condition we had on performatively stable points:
    the optimal model for the current distribution is the model we deployed to get this distribution.

    In a reinforcement learning setting, one would define a regret and then try to find a policy with minimal regret.
    However, \citet{brown2022performative}'s definition of the optimal strategy for stateful performative prediction
    is based on the idea of deploying a single model (i.e., performing a single \emph{action} in the RL lens)
    that is optimal in terms of the \emph{long-term} loss.
    To define the long-term loss of a model,
    we first define the \emph{limiting distribution map} for a fixed model \(\theta\) as
    \begin{align}
        \mathcal{D}_\infty (\theta) := \lim_{t\to\infty} Q _t &&\text{where } Q_t = \mathrm{Tr}(\theta, Q _{t-1})
        \label{eq:limiting-distribution-map}
    \end{align}
    and where \(Q_0\) is an arbitrary initial distribution.
    To get \(\mathcal{D}_\infty\), we deploy the same classifier at every time step.
    The conditions under which the limit converges and is unique (independent of the initial \(Q_0\))
    can be found in claim 1 in \citet{brown2022performative}.

    We define then the \emph{long-term performative risk}~\cite{izzo2022learn} as
    \begin{align}
        \mathit{PR}_\infty (\theta) := \mathbb{E} _{z\sim Q^*(\theta)}\big[\ell(z; \theta)\big]~,
        \label{eq:long-term-perf-risk}
    \end{align}
    i.e., the expected loss on the limiting distribution map of \(\theta\).
    The \emph{long-term performatively optimal point}, \(\theta_\mathit{SPO}\), is the model that minimizes \eqref{eq:long-term-perf-risk}:
    \begin{align}
        \theta_\mathit{SPO} := \argmin_{\theta\in \Theta}\mathit{PR}_\infty (\theta)~.
        \label{eq:long-term-perf-optimal}
    \end{align}

% \thomas{Move somewhere else: Other works with this setup, or a similar one, are \citet{wood2021online} and \citet{li2022state}. (neither of these papers is about the stateful setting)}
%rare in the performative prediction literature and too close to reinforcement learning to be interesting to us.

%%% Local Variables:
%%% mode: LaTeX
%%% TeX-master: "../main"
%%% End:
    % Section 2: Background
\section{The distribution map}%
\label{sec:distribution-map}
The data requirements for experiments on performative predictions are very different from those for the usual ML experiments.
In a typical ML setup, there is a training distribution \(\gD_\mathit{train}\) and a test distribution \(\gD_\mathit{test}\),
but in the case of performative predictions, we need a distribution \emph{map},
i.e., a function returning a distinct distribution for different argument values (see \eqref{eq:distribution-map}).
 
One possible approach is to collect the distribution map from real-world experiments
where new models \(\theta_i\) are deployed and the induced distribution \(\gD_i\) recorded.
However, this approach has several weaknesses:
(i) the distribution map will only be defined for those \(\theta_i\) that were actually tried,
which likely precludes finding performatively optimal or stable points;
(ii) the distribution map, \(\gD(\theta)\), will rely on a specific interpretation of \(\theta\), which means we cannot,
for example, change the model architecture afterwards, because the interpretation of \(\theta\) has already been locked in;
(iii) depending on the nature of the data, recording data this way could be very expensive or even unethical.
These weaknesses make it very difficult to collect a real-world dataset. Insofar as we are aware, no existing works have taken this path.
% This approach would retrieve samples which are very expensive to obtain (level 1a access to the distribution map).  

Another approach is to assume a compact mathematical model for the distribution map.
This can be done in two main ways:
In the first option, we base the distribution map on distribution in parametric families,
e.g., a mixture of Gaussians where the locations depend on \(\theta\)~\cite{izzo2021learn,miller2021outside}.
This results in a purely synthetic dataset.
The second option is to use a semi-synthetic approach where we use a real, static dataset as the base
and we augment this dataset with a mathematical mechanism to add the performative aspect,
i.e.\ we have a base distribution and define a transformation of the old sample that depends on \(\theta\).

Note that even if we have an experimental framework with a full mathematical model of the distribution map,
we may still want to only use it as a black box for drawing samples, in order to simulate a situation where only samples are available.

The structure of this section is as follows:
To dissect the distribution map, we begin by discussing common mathematical models of distribution maps.
Afterwards, we present classification schemes for distribution maps. Finally, we present common datasets used in experiments for performative prediction.

\subsection{Mathematical models for distribution maps}%
\label{ssec:mathematical-models-for-distribution-maps}

\subsubsection{Arbitrary mechanisms specified in functional form}%
\label{sssec:arbitrary-mechanisms-specified-in-functional-form}
The simplest mathematical model of a distribution map is one which is specified in terms of well-known parametric distributions.
For example, consider the \emph{pricing} dataset from~\citet{izzo2021learn}:
it is defined as \(\mathcal{D}(\theta)=\mathcal{N}(\mu_0-\epsilon\theta; \Sigma)\),
where \(\mathcal{D}(\cdot;\cdot)\) is the Gaussian distribution,
and \(\mu_0\), \(\epsilon\) and \(\Sigma\) are constants.

\subsubsection{Strategic classification}%
\label{sssec:strategic-classification}
Arguably the most commonly considered mechanism for performative prediction is \emph{strategic classification}~\cite{hardt2016strategic}.
In this model, a user intends to alter a classification from a negative to a positive outcome by changing their features,
but at the same time, the user is restricted in the kinds of changes they can make.

For example, when we consider our bank loan example,
the user would like to get a positive loan decision
and can achieve this by changing the financial features that the bank uses for the decision.
However, such feature values can, of course, not be changed arbitrarily.
There is a cost associated with changing feature values; and some features (like date of birth) cannot be changed at all
or can only be changed in one direction (like the number of previous loans).

% Strategic classification is a model that predates performative prediction.
%

% When a user is incentivized to alter the model's prediction from a negative to a positive outcome,
% performative prediction is an instance of strategic classification.
%
% An example is college admissions.
%
% If a student is rejected by a university, they have the opportunity to modify certain features
% (e.g., improve standardized test scores or secure better recommendation letters) and reapply the following year.

The mathematical formalization of strategic classification is as follows:
The original features of the users are captured by a  base distribution, \(\gD_\mathrm{base}\),
over feature vector and label tuples.
The features then change such that they 
maximize a utility function, \(u:\gX\times\Theta\to \R\), while minimizing a cost function, \(c:\gX\times \gX\to \R\):
\begin{equation}
  x = \argmax_{x'\in \gX}\;\big(u(x', \theta)- c(x', x_\mathrm{base})\big),\quad y = y_\mathrm{base} 
  \quad \text{ where }(x_\mathrm{base}, y_\mathrm{base})\sim \gD_\mathrm{base}
  \label{eq:strategic-classification}
\end{equation}
Intuitively, this transformation describes a user with a goal that is encoded by the utility function, \(u\) (usually the goal is to get a high score with features \(x'\) on the model defined by \(\theta\)) and with a personal cost, \(c\), of changing their features from \(x_\mathrm{base}\) to \(x'\), normally a distance.

In the loan example, the utility function would be binary:
returning 1 for an accepted loan and 0 otherwise.
The loan applicant thus aims to modify their features such that the loan is accepted.
Those feature values that cannot be changed get an infinite cost for changing them.
For features that can be changed, it is often assumed that the cost grows quadratically in the difference of the feature values.
This means, it is not too difficult to change the feature by a small amount,
but large changes quickly become infeasible.
%
% DELETION CANDIDATE
This matches the experience that increasing your income, for example, might be possible to a small degree --- e.g., by getting a second job ---
but becomes very hard to do to a larger degree.

% This is the case in most of the performative prediction papers that use strategic classification. 

It is evident then, that the new distribution of $(x, y) \sim \mathcal{D}(\theta)$ would follow the same distribution as $\mathcal{D}_{base}$
except that the features have been modified according to \eqref{eq:strategic-classification} in a way that depends on $\theta$.
Strategic classification therefore defines a distribution map as used in performative prediction.

%\javi{In general, ``traditional'' strategic classification works focus on how to train a model that is robust to these shifts in the input (citation needed). The problem transforms then into minimising the loss for the samples found in equation \ref{eq:strategic-classification}.} %The naive approach is to train the model with the datapoints obtained from \ref{eq:strategic-classification} (citation needed). Although others.... }

%\javi{Talk about training, how as dataset distr is known, you can optimize directly with a min-max game}

\subsubsection{Resampled-if-rejected procedure}%
\label{sssec:resampled-if-rejected-procedure}

One problem with applying a known transformation to individual samples of a real dataset (as is done in strategic classification) is that the sample might not be realistic after the transformation.
To address this problem, \citet{mofakhami2023performative} define a new transformation in the bank loan example. 

First, the authors propose that users will only transform their feature values if they are rejected when applying for a loan.
The specific mechanism by which they transform the feature values is then this:
users randomly swap the values of some of their features for the values from other users.
%
%Note that this data shift uses another notion of strategically gaming the model different from \textit{strategic classification}.
In other words, the users no longer try to improve their score by explicitly and directly maximizing their score --- as they do in \emph{strategic classification} --- but rather by \emph{mimicking} other randomly-chosen users.

Let \(p(z)\) be the probability of encountering the sample \(z\) in the base distribution.
Then we can express \(p_\theta(z)\) --- the probability of encountering the sample \(z\) after deploying the classifier with model parameters \(\theta\) --- as:
\begin{align}
    p_\theta(z)=p(z)\big(1-g_\theta(z)+C_\theta\big)
    \label{eq:resampled-if-rejected}
\end{align}
where \(g_\theta(z)\) is the probability for rejection, and $C_\theta=\mathbb{E}_{z\sim p(\cdot)}\big[g_\theta(z)\big]$ is the expectation value of \(g_\theta\) over \(p(z)\).
What \eqref{eq:resampled-if-rejected} shows is that the probability of encountering \(z\) is reduced by the chance that \(z\) is rejected, \(g_\theta(z)\).
The expectation value $C_\theta$ is then added to the probability in order to ensure that the probability distribution remains normalized.
Intuitively, the added $C_\theta$ corresponds to the chance that a \emph{different} sample is rejected and $z$ is chosen instead.
In other words, for any sample, there is a constant probability of $C_\theta$ that this sample is chosen after a rejection.

\subsubsection{MNIST with groups}%
\label{sssec:mnist-with-groups}

Aiming to study group fairness, \citet{jin2024addressing} consider cases where, instead of directly modifying samples in the data, the deployed model causes a shift of the proportions of groups in the dataset.

Let the distribution induced by the distribution map be a mixture of two distribution maps: $\mathcal{D}_1$ and $\mathcal{D}_2$, each representing a group.
Each distribution has a weight factor, $p^t_1$ and $p^t_2$ respectively, determining what proportion of overall samples come from each distribution, with $p^t_1+p^t_2=1$.
Upon model deployment, the weights, $p^t_1(\theta)$, $p^t_2(\theta)$, change but the distributions $\mathcal{D}_1$, $\mathcal{D}_2$ remain unchanged.
This simulates a scenario where the user base stems from two groups of people and the performativity effect changes the proportion of the groups in the user base, but leaves the users otherwise unchanged.

\subsection{Classification of distribution maps}%
\label{ssec:classification-of-distribution-maps}
Especially for theoretical work, it is very useful to discuss certain families of distribution maps that satisfy some constraints.

\begin{figure}
    \centering
    \begin{subfigure}[b]{0.42\textwidth}
        \includegraphics[width=\linewidth]{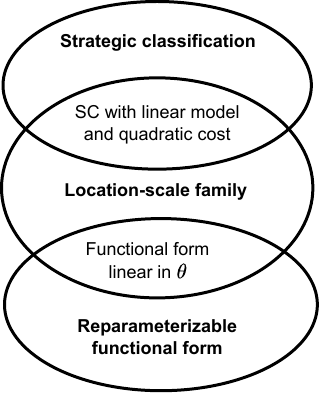}
        \caption{%
            Venn diagram showing the overlap of different sub-families of the base-distribution family.
            ~\\
        }%
        \label{fig:venn-diagram-inner}
    \end{subfigure}
    \begin{subfigure}[b]{0.78\textwidth}
        \centering
        \begin{tabular}{lll}
            \toprule
            \textbf{Sub-families of BDF} & \textbf{Example} & \textbf{Ref} \\
            \toprule
            \textbf{Strategic}           & \(z_ \mathrm{base}\sim \mathcal{D}_ \mathrm{base}\) & \cite{perdomo2020} \\
            \textbf{classification}      & \(\varphi(z_ \mathrm{base}; \theta)\) \\
                                         & \(\quad=\argmax_{z'} u(z', \theta) - c(z', z_\mathrm{base})\) \\
            \midrule
            \textbf{Location-scale}      & \(z_ \mathrm{base}\sim \mathcal{D}_ \mathrm{base}\) & \cite{miller2021outside} \\
            \textbf{family}              & \(\varphi(z_\mathrm{base}; \theta)\) \\
                                         & \(\quad= (\Sigma_0+\Sigma(\theta))z_\mathrm{base} + \mu_0 + \mu(\theta)\) \\
            \midrule
            \textbf{Reparameterizable}   & \(z_ \mathrm{base}\sim \mathcal{N}(0; 1)\) & \cite{izzo2021learn} \\
            \textbf{functional form}     & \(\varphi(z_ \mathrm{base}; \theta) = z_ \mathrm{base} + \sqrt{a_1\theta+a_0}\) \\
            \midrule
            % Base-distribution family     & \(z=\varphi(z_\mathrm{base}; \theta), z_\mathrm{base}\sim \mathcal{D}_\mathrm{base}\) \\ \midrule
            Strategic classification     & \(z_ \mathrm{base}\sim \mathcal{D}_ \mathrm{base}\) & \cite{perdomo2020} \\
            with linear model            & \(u(z, \theta)=\theta^T z +\mu\) \\
            and quadratic cost           & \(c(z, z') = \tfrac12\varepsilon\|z-z'\|_2^2\) \\
                                         & \(\varphi(z_\mathrm{base}; \theta)=z_\mathrm{base} + \varepsilon^{-1}\theta\) \\
            \midrule
            % Strategic classification     & \(\varphi(z_\mathrm{base}; \theta)\) \\ 
            %                              & \(=\argmax_{z'} u(z', \theta) - c(z', z_\mathrm{base})\) \\ \midrule
            % Functional form              & \(z\sim \mathrm{Poisson}(\lambda)\) \\ \midrule
            Functional form which        & \(z_ \mathrm{base}\sim \mathcal{N}(0; 1)\) & \cite{izzo2021learn} \\ 
            is linear in \(\theta\)      & \(\varphi(z_ \mathrm{base}; \theta) = \sigma z_ \mathrm{base} + \mu\theta\) \\
            \bottomrule
        \end{tabular}
        \caption{%
          Examples for each of the regions in the Venn diagram.
          The references point to the source of the examples.
        }%
        \label{tab:venn-diagram-examples}
    \end{subfigure}
    \caption{%
            A diagram and a table of different sub-families of the \emph{base-distribution family} (Section~\ref{ssec:base-distribution-with-model-dependent-transformation}):
            \emph{Strategic classification} (Section~\ref{sssec:strategic-classification}),
            \emph{location-scale family} (Section~\ref{sssec:location-scale-families}),
            and (reparameterizable) \emph{functional forms} (Section~\ref{sssec:arbitrary-mechanisms-specified-in-functional-form}).
            The Venn diagram on the left shows how these sub-families can overlap
            and on the right, there are examples for each distinct region in the Venn diagram.
            An overlap of the first circle (strategic classification) and the third circle (reparameterizable functional form) is also conceivable --- for example a strategic classification setup with a synthetic base distribution --- but does not represent an instructive category.
    }%
    \label{fig:venn-diagram}
    % Description of the figure for visually-impaired readers:
\end{figure}

% maybe emphasize more here what a natural fit the base-distribution family is for PP (base distribution that is shifted by performativity)
\subsubsection{The base-distribution family}%
\label{ssec:base-distribution-with-model-dependent-transformation}
The base-distribution family is one of the least constrained families of distribution maps
and thus encompasses almost all of the distribution maps that are considered in the literature.
\citet{cyffers2024optimal} call it \textit{push-forward} model but we use the term \textit{base distribution} in analogy to the location-scale family (Section~\ref{sssec:location-scale-families}) from earlier work.
Any distribution map that is based on strategic classification (Section~\ref{sssec:strategic-classification}) is, for example, a member of this family.

Formally, a member of this family is defined by a tuple \((\gD_\mathrm{base}, \tf)\) consisting of
a base distribution, \(\gD_\mathrm{base}\in \Delta(\gZ)\), and a transformation function, \(\tf: (\gZ\times \Theta)\to \gZ\),
which transforms individual samples.
In order to sample from the distribution map \(z\sim\gD(\theta)\),
we first sample from \(\gD_\mathrm{base}\) and then transform with \(\tf\), which takes \(\theta\) as an argument:
\begin{equation}
  z = \tf(z_\mathrm{base}, \theta)\quad\quad\text{where }~ z_\mathrm{base}\sim \gD_\mathrm{base}~.
  \label{eq:dist-map-with-base-dist}
\end{equation}
This results in a distribution map whose dependence on \(\theta\) is entirely contained in the transformation function, $\varphi$.

This pattern of a \emph{base distribution} and a \(\theta\)-dependent transformation
is, however, not just a mathematical trick.
Many occurrences of performative prediction in the world are naturally thought of
as consisting of a ``base case'', where no performativity has taken place yet,
and a mechanism which transforms this base case.
For example, in the credit scenario, applicants have their ordinary, unoptimized features,
\(x_\mathrm{base}\), which were collected before they acted strategically in any way,
and which they then \emph{modify} in response to the classifier, \(\theta\).
This can be understood as a \(\theta\)-dependent transformation.
Similarly, in the situations where a forecast influences behaviour,
there is a base distribution of behaviour that would have happened if the forecast had not existed,
and only the existence of the forecast changes that base distribution.

This \emph{base-distribution family}
has previously appeared in, e.g., \citet{miller2021outside,mofakhami2023performative}.
Many sub-families of the base-distribution family exist,
including \emph{location-scale families}, which are discussed in the next section.
Fig.~\ref{fig:venn-diagram} shows how the sub-families of the base-distribution family intersect.

% Of course, any distribution map \(\gD(\cdot)\) can be reformulated in this way,
% if we allow arbitrarily complex transformation functions \(\tf\),
% but the point is that the base distribution \(\gD_\mathrm{base}\) often has a very natural interpretation
% as the ``default state'' in which no performativity has taken place yet.
%
% This is the case in most of the motivating examples.

\subsubsection{Location(-scale) families}%
\label{sssec:location-scale-families}
The base-distribution family appears in both semi-synthetic and synthetic datasets (parametric forms),
but for synthetic datasets, we can restrict the form of the distribution map even further,
which allows us to guarantee certain theoretical properties, like mixture dominance
(discussed in Section~\ref{sssec:convex-optimization-with-location-scale-families}).

For example, \citet{miller2021outside} restricted the distribution map to \emph{location-scale families}.
A location-scale family is a family of distributions
that is parameterized by a location \(\mu\in\mathbb{R}^d\) and a scale \(\Sigma\in\mathbb{R}^{d\times d}\).
For each family, there is a base distribution, which we may call \(\mathcal{D}_ \mathrm{base}\),
and we can sample \(z\sim P_{\mu, \Sigma}\) as follows:
\begin{equation}
    z = \Sigma z_ \mathrm{base} + \mu\quad\quad\text{where }z_ \mathrm{base}\sim \mathcal{D}_ \mathrm{base}~.
    \label{eq:location-scale-simple}
\end{equation}
In order to introduce a dependence on the model parameters, \(\theta\),
we make the location and the scale depend on \(\theta\):
\begin{equation}
    z = \big(\Sigma_0 + \Sigma(\theta)\big)z_ \mathrm{base} + \mu_0 + \mu(\theta)
    \label{eq:location-scale-theta}
\end{equation}
where \(\Sigma(\cdot)\) and \(\mu(\cdot)\) are linear functions in \(\theta\).
Comparing with~\eqref{eq:dist-map-with-base-dist},
we can see that assuming location-scale families means restricting the form of the transformation function \(\tf\).

Later works, like~\cite{lin2024plugin}, restricted the transformation function even further,
to \emph{location} families:
\begin{equation}
    z = z_ \mathrm{base} + \mu(\theta)
    \label{eq:location-family}
\end{equation}
where \(\mu(\cdot)\) is again a linear function in \(\theta\).

The distribution map generated by \emph{strategic classification} has this form
if the cost function is quadratic and the score function is given by a linear model
(see Section~\ref{sssec:strategic-classification}).

\subsubsection{Outcome performativity}\label{sssec:outcome-perf-def}
    We have mentioned \emph{outcome performativity} as a special case of performative prediction previously,
    but we have not precisely defined it so far.
    Let us remedy this situation here.
    The structure of the distribution map in outcome performativity corresponds to a base-distribution family,
    with some caveats that we will see.
    Let us consider a simplified version of outcome performativity first.
    Let \(\mathcal{D}^X_ \mathrm{base}\) be a base distribution over a space \(\mathcal{X}\).
    and let \(\Gamma: \mathcal{X}\times \mathcal{Y}\to \mathcal{Y}\) be a transformation.
    To sample \((x,y)\sim \mathcal{D}(\theta)\), we perform the following operation
    \begin{align}
      x=x_ \mathrm{base},\quad
        y=\Gamma\big(x_ \mathrm{base}, f_ \theta(x_ \mathrm{base})\big) && \text{where } x_ \mathrm{base}\sim\mathcal{D}^X_ \mathrm{base}
    \end{align}
    and where \(f_\theta\) is our model.
    There are four things of note here:
    (i) \(x\) is not transformed at all, meaning that the marginal distribution over \(x\) is not affected by the distribution shift;
    (ii) there is no \(y\) in the base distribution;
    and (iii) the model weights \(\theta\) affect the transformation only through the model's output on \(x_ \mathrm{base}\)
    (or, equivalently, \(x\)),
    meaning that all performativity goes through \(\hat y = f_ \theta(x)\).
    It is primarily the properties (i) and (iii) which make this \emph{outcome} performativity:
    it is only the outcome, \(y\), which is transformed and the transformation only depends on what we predicted the outcome to be, \(\hat y\).
    Property (iii) additionally describes a kind of \emph{locality}
    because the distribution shift on a data point \(x\) only depends the part of the model \(\theta\) that affects that point \(x\) ---
    any change of the model that only affects other inputs has no effect on the \(y\) that will be assigned to \(x\).

    The non-simplified version of outcome performativity additionally allows randomness in the transformation.
    We model this the following way:
    let the base distribution \(\mathcal{D}_ \mathrm{base}\) be over the Cartesian product of \(\mathcal{X}\) and a sufficiently large domain, \(\mathcal{U}_y\) for a random variable, e.g., \(\mathcal{U}_y=\mathbb{R}\),
    with the caveat that we never observe samples from \(\mathcal{U}_y\) directly.
    Let also \(\Gamma': \mathcal{X}\times \mathcal{Y} \times \mathcal{U}_y\to \mathcal{Y}\).
    We have then~\cite{kim2023making}
    \begin{align}
        x=x_ \mathrm{base},\quad
        y=\Gamma'\big(x_ \mathrm{base}, f_ \theta(x_ \mathrm{base}), \xi_y\big)  && \text{where } (x_ \mathrm{base}, \xi_y)\sim\mathcal{D}_ \mathrm{base}~.
    \end{align}
    In this formulation, \(\Gamma'\) remains a deterministic function and \(\xi_y\in \mathcal{U}_y\) is a nuisance variable providing the randomness.
    The fact that the distribution over \(\mathcal{U}_y\) cannot be observed will require special handling when we fit a model to distribution map observations.
    The graduation rate intervention example given in Section~\ref{ssec:guiding-examples} is an instance of outcome performativity:
    the eventual outcome, \(y\) (graduation), depends on the original features \(x\) (past academic performance)
    and on whether the students received additional help, which was determined by the original prediction, \(\hat y\)
    (whether they are likely to graduate on the current course).
    The example has the locality property mentioned above as long as the school district has enough resources
    such that academic support assigned to one student does not diminish the support that other students receive.
    The car navigation example is not an instance of outcome performativity --- at least not straightforwardly so.
    For one, the true outcome, the final travel time, very much depends on the recommendations that the other drivers received,
    which breaks locality.

\subsubsection{Transition maps from distribution maps}%
\label{sssec:transition-maps-from-distribution-maps}
    Transition maps, \(\mathrm{Tr}(\theta, Q _{t-1})\), play the role of distribution maps \emph{stateful} performative prediction
    (see Section~\ref{ssec:stateful-performative-prediction} for the definition).
    In the following, let \(Q_t\in\Delta(\mathcal{Z})\) be the distribution observed at a time \(t\).
    Given a distribution map \(\mathcal{D}(\theta)\),
    there are several simple ways to define a transition map on top of them.

    Imagine that only a subset of applicants respond to the deployed model at each time step.
    That is, only a certain proportion of the environment goes through the distribution shift
    while the remaining fraction, \(\lambda\in [0;1]\), stays as it was.
    We can model this as
    \begin{align}
        \mathrm{Tr}(\theta_t, Q _{t-1} ) := \lambda Q _{t-1} + (1-\lambda) \cdot \mathcal{D}(\theta_t)
        \label{eq:geometric-decay}
    \end{align}
    where \(Q _{t-1}\) is the distribution from the previous time step
    and the ``\(+\)'' indicates a mixture of the two distributions.
    This kind of transition map has been referred to as a \emph{geometrically decaying environment}.
    It should be straight-forward to see that this transition map converges to a \emph{limiting distribution map}
    when the same model \(\theta\) is deployed at every time step,
    as described in Section~\ref{ssec:stateful-performative-prediction}.

    Another way to to model the basic idea of ``different parts of the environment respond at different rates'' is to split the environment into a uniform mixture of \(k\) distributions and to let the first component of the mixture respond to the latest deployed model, while the second component responds to the model before that, and so on~\cite{brown2022performative}.
    Formally, at each time step \(t\), we have \(k\) distributions \(Q_t^j\), \(j=1\dots k\) such that
    \begin{align}
        Q_t^j = \begin{cases}
            \mathcal{D}(\theta_{t-j+1}) &\text{if }t-j+1\geq 1\\
            Q_0^j &\text{otherwise}
        \end{cases}
    \end{align}
    where \(Q_0^j\) is the initial distribution for component \(j\).

\subsection{Fitting a mathematical model to distribution map observations}%
\label{ssec:distribution-map-model-fitting}
    Many of the methods that we will discuss in Sections~\ref{sec:opt-perf-stable} and \ref{sec:opt-perf-opt}
    require a mathematical model of the distribution map,
    in order to, for example, compute gradients through the distribution map (see e.g.\ Section~\ref{ssec:model-based}).
    In the experimental setup, we typically have either a full mathematical model of the distribution map
    or at least a mathematical model of the performative mechanism
    (e.g., the transformation function in the base-distribution family).
    in the case of semi-synthetic data.
    These mathematical models could be made available to the method we wish to utilize,
    but that would be a highly unrealistic setup:
    in a real deployment, we would only observe \emph{samples} and would not have a mathematical model of the distribution map.

    To make the setup more realistic,
    we can fit a model to samples from the distribution map.
    This requires, however, that a reasonable hypothesis class is known for the distribution map model.
    % We discuss the details in the following.
    % Note that we can either learn a full mathematical model of the distribution map
    % or just the performative mechanism.

\subsubsection{Levels of access}%
\label{sssec:levels-of-access}

    % In practice, option (ii) requires fitting a mathematical model of the distribution map to observed samples (as described in Section~\ref{ssec:distribution-map-model-fitting}).
    Depending on the method used, there are different requirements for the distribution map model.
    We can distinguish two levels of requirements:
    (i) requiring a full model of the probability densities of the distributions returned by the distribution map;
    (ii) only requiring a model of the \emph{dynamics} (i.e., transformation function) of a distribution from the base-distribution family.

    We combine this distinction of different levels of distribution map models
    with the distinction previously introduced in Section~\ref{sssec:distribution-map-known}
    for the differences in sample access (cheap vs expensive samples),
    in order to create a combined hierarchy of \emph{levels of access to the distribution map}:
\begin{itemize}
    \item \textbf{Level 1: black-box sample access to the distribution map}:
        \begin{itemize}
            \item \textbf{1a: Expensive sample access after real deployment} (used by methods discussed in Section~\ref{ssec:model-free}):
                we only get a limited amount of samples and need to build our model from that, or use Bandit methods.
            \item \textbf{1b: Cheap sample access through black-box simulation}
                (used by methods discussed in Section~\ref{ssec:repeated-risk-minimisation}):
                we can draw an unlimited number of (simulated) samples from the distribution map,
                but do not have white-box access.
        \end{itemize}
    \item \textbf{Level 2: mathematical model of the distribution map}:
        \begin{itemize}
            \item
            \textbf{2a: Parameterized model of the transformation function of a base-distribu\-tion family} (used by methods discussed in Section~\ref{sssec:convex-optimization-with-location-scale-families}):
                we are able to split the distribution map into a static (base distribution) and a dynamic part (transformation function).
                We have a mathematical model of the dynamic part and access to samples from the static part.
                % and have collected data for the static part.
                %
            \item
            % \thomas{parameterized model transformation function + parameterized model of base distribution}
            \textbf{2b: Parameterized model of the probability density of the distribution map output} (used by methods discussed in Section~\ref{sssec:generic-differentiable-model-of-the-distribution-map}):
                we have a mathematical model
                % (that may have some free parameters)
                of the probability density of the image of the distribution map and can optimize it directly.
                In case of a distribution map from the base-distribution family,
                this requires a model of the dynamic part (the transformation function) \emph{and} the static part (a parameterized model of the base distribution).
        \end{itemize}
\end{itemize}
    We can use this hierarchy to classify the methods presented in Sections~\ref{sec:opt-perf-stable} and \ref{sec:opt-perf-opt},
    which vary in terms of the required level of access.
    The techniques discussed in the following section can usually be used for learning both 2a and 2b distribution map models.
    It all depends on the hypothesis class:
    In order to learn a 2a model, a hypothesis class over \emph{transformation functions} is sufficient;
    for a 2b model, one needs a hypothesis class over probability densities,
    which can, however, still be parameterized as a base distribution family with a parameterized base distribution and a parameterized transformation function.

\subsubsection{General techniques}

\citet{lin2024plugin} coined the term \emph{distribution atlas} for a parameterized collection of distribution map models: \(\bm{\mathcal{D}}_\mathcal{B}\coloneqq\{\mathcal{D}_\beta\}_{\beta\in\mathcal{B}}\), where \(\mathcal{B}\subset \mathbb{R}^d\) is a parameter space.
Based on observed samples, we determine the value of \(\beta\) and thus the distribution map model to be used in the optimization.
This process can be formalized~\cite{lin2024plugin} with a function \(\widehat{\mathrm{Map}}\), which estimates \(\beta\) from samples:
\begin{equation}
    \hat{\beta} = \widehat{\mathrm{Map}}\big((\theta_1, z_1), \dots, (\theta_n, z_n)\big)~,
\end{equation}
such that we can estimate the performative risk as
\begin{equation}
    \mathit{PR}^{\hat{\beta}}(\theta)=\mathbb{E}_{z\sim\mathcal{D}_{\hat\beta}(\theta)}\big[\ell(z;\theta)\big]~.
\end{equation}
The use of a fitted distribution map model incurs two kinds of errors~\cite{lin2024plugin}:
the statistical error from imperfect fitting from finite samples
and the misspecification error from the fact that the hypothesis class does not contain the real model.
If \(\beta^*\) refers to the optimal choice of \(\beta\) within the hypothesis class, we can define the two errors as:
\begin{align}
    \mathrm{StatErr} &= \sup_{\theta\in\Theta}\left|\mathit{PR}^{\beta^*}(\theta)-\mathit{PR}^{\hat{\beta}}(\theta)\right|\\
    \mathrm{MisspecErr} &= \sup_{\theta\in\Theta}\left|\mathit{PR}^{\beta^*}(\theta)-\mathit{PR}(\theta)\right|
\end{align}
Despite these errors that are incurred by a fitted distribution map model,
\citet{lin2024plugin} argue that the benefits of using efficient optimization still outweigh the downsides of the error.
The authors propose the following protocol:
\begin{enumerate}
    \item
        Use an exploration distribution over models, \(\theta\), to collect samples from the distribution map.
    \item
        Use the collected observations, \({\{(\theta_i, z_i)\}}_{i=1}^N\),
        to fit an estimate of the distribution map: \(\mathcal{D}_{\hat\beta}(\cdot)\).
    \item
        Optimize the performative risk on \(\mathcal{D}_{\hat\beta}(\cdot)\) in order to get \(\widehat{\theta}_\mathit{PO}\), the optimal point according to the estimated distribution map.
\end{enumerate}
The authors prove different bounds on the excess risk, \(|PR(\widehat{\theta}_ \mathit{PO})-PR(\theta_ \mathit{PO})|\),
which depend, among other things, on the regularity of the loss and the misspecification.

Another approach for addressing misspecification error was given by~\citet{xue2024distributionally},
who use the framework of distributionally robust optimization (DRO).
The basic idea in distributionally robust optimization is to optimize for the worst case
within a distribution over possible datasets.
In the case of performative prediction, we consider a set of possible distribution maps
around our best guess of the distribution map, and optimize for the worst case within this set.
Specifically, the paper defines an uncertainty collection, \(\mathcal{U}\), around a given distribution map, \(\mathcal{D}\):
\begin{equation}
    \mathcal{U}(\mathcal{D}) = \Big\{\widetilde{\mathcal{D}} : \Theta\to \Delta(\mathcal{Z})\,\Big|\,
    \mathit{KL}\big(\widetilde{\mathcal{D}}(\theta)\|\mathcal{D}(\theta)\big) \leq \rho, \forall\theta\in\Theta
    \Big\}
    \label{eq:uncertainty-collection}
\end{equation}
where \(\mathit{KL}(\cdot\|\cdot)\) refers to the Kullback–Leibler divergence.
Based on this, we define the \emph{distributionally robust performative risk}:
\begin{equation}
    DRPR(\theta) = \sup_{\widetilde{\mathcal{D}}\in \mathcal{U}(\mathcal{D})}
    \mathbb{E}_{z\sim \widetilde{\mathcal{D}}(\theta)}\big[\ell(z; \theta)\big]~.
    \label{eq:dro-perf-risk}
\end{equation}
The paper's theoretical analysis and experiments show that even if the misspecification is large,
optimizing DRPR can still ensure a low \emph{true} performative risk.

\subsubsection{Techniques for outcome performativity}\label{sssec:learn-outcome-performativity}
    We discuss here separately the case where the distribution map corresponds to the \emph{outcome performativity} setting,
    which represents a much easier case.
    As described in Section~\ref{sssec:outcome-perf-def},
    in this case the distribution map is fully determined by a transformation function \(\Gamma: \mathcal{X}\times \mathcal{Y}\times \mathcal{U}_y\to \mathcal{Y}\),
    where \(\mathcal{X}\) is the feature space, \(\mathcal{Y}\) is the label space and \(\mathcal{U}_y\) is the domain of a nuisance variable, \(\xi_y\).
    However, instead of learning the transformation function \(\Gamma\) and the distribution over \(\mathcal{U}_y\) separately,
    we can directly learn a representation of the conditional probability distribution, \(P(y|\hat y, x)\),
    which --- together with the base distribution over \(\mathcal{X}\) --- contains all required information, as can easily be shown:
    Let \(p_\theta = \mathcal{D}(\theta)\) be the probability density returned by the distribution map for \(\theta\).
    We can express \(p_\theta\) in terms of in terms of the base density, \(p_ \mathrm{base}\), corresponding to \(\mathcal{D}^X_ \mathrm{base}\), the base distribution over \(\mathcal{X}\) and the conditional probability distribution:
    \begin{align}
        p_ \theta(x, y) := p_ \mathrm{base}(x)\, P\big(y|\hat y=f_\theta(x), x\big)~.
        \label{eq:outcome-perf-pi-dist-map}
    \end{align}
    Thus, the conditional distribution, \(P(y|\hat y, x)\), and \(p_ \mathrm{base}(\cdot)\) together fully specify \(p_\theta\).
    % the base distribution, \(\mathcal{D}_ \mathrm{base}\),
    % together fully specify the distribution map in the case of outcome performativity.

%
    % Additionally, the distribution over \(\mathcal{U}_y\) is unobservable and needs to be learned.
    % However, our task is easier if instead of learning the transformation \(\Gamma\) and the distribution over \(\mathcal{U}_y\),
    % we learn a function \(q(x)\), which directly returns the conditional probability distribution \(P(y|\hat y, x)\),
    % where \(x\) is fixed.
    As both \(y\) and \(\hat y\) are discrete variables,
    we can parameterize the conditional probability distribution
    as a function from \(\mathcal{X}\) to a matrix of size \(|\mathcal{Y}|\cdot|\mathcal{Y}|\),
    where each entry corresponds to a possible pair \((y, \hat y)\).
    So, let \(q(x)\) be the function representing \(P(y|\hat y, x)\),
    then for any \(x\), \(M=q(x)\) is a matrix with \(M_{ij} = P(y=i|\hat y=j, x)\).
    In the binary case, where \(y, \hat y \in\{0, 1\}=\mathcal{Y}\), \(q(x)\) only needs to return a vector with two entries in the interval \([0, 1]\):
    \begin{align}
        q: \mathcal{X}\to {[0, 1]}^2, \quad q(x) := \begin{pmatrix}P(y=1|\hat y = 0, x)\\ P(y=1|\hat y = 1, x)\end{pmatrix}
    \end{align}
    because we have \(P(y=0|\hat y, x)=1-P(y=1|\hat y, x)\).
    % With \(q(\cdot)\), we can define a function \(\pi_y: \mathcal{X}\times \mathcal{Y} \to \Delta(\mathcal{Y})\),
    % which returns the conditional probability distribution, such that \(y\sim \pi_y(x, \hat y)\) with \(P(y|\hat y, x)\),
    % assuming the binary case:
    % %
    % \begin{align}
    %   \pi_y(x, \hat y) = \mathrm{Bernoulli}(p)&&\text{where }p =
    %     \begin{cases}
    %       q(x_ \mathrm{base})_1 & \text{if }\hat y=0\\
    %       q(x_ \mathrm{base})_2 & \text{otherwise}~.
    %     \end{cases}
    %     \label{eq:outcome-perf-q-pi}
    % \end{align}
    % %
    % In other words, \(y\) is sampled from a Bernoulli distribution whose probability \(p\) is given by \(q(x)\).
    % The distribution map, \(\mathcal{D}(\theta)\) is then
    % %
    % \begin{align}
    %   x=x_ \mathrm{base}, \quad y\sim \pi_y(x_ \mathrm{base}, f_\theta(x_ \mathrm{base})), &&\text{where }~ &x_ \mathrm{base}\sim \mathcal{D}_ \mathrm{base}~.
    %     \label{eq:outcome-perf-pi-dist-map}
    % \end{align}
    %

%
    The question of when \(q(\cdot)\) can be learned from observed triplets \((x, \hat y, y)\) was investigated by \citet{mendler2022anticipating} and \citet{kim2023making}.
    The main challenge is that if the \(\hat y\)'s in all the triplets were generated by a single model \(\hat y=f(x)\),
    then \(q(\cdot)\) will learn \(P(y|\hat y=f(x'), x=x')\) and will not learn to model the real effect of \(\hat y\) on \(y\).
    In other words, the data generated by a single model lacks coverage.
    The authors of the cited works nevertheless identify situations in which \(q(\cdot)\) can be learned.
    The concrete learning algorithm for \(q(\cdot)\) (and thus \(\pi_y\)) that is proposed in \citet{kim2023making}%,
    called \emph{POI-Boost},
    is based on the concept of \emph{outcome indistinguishability}~\cite{dwork2021outcome,gopalan2023loss},
    which ensures that \(\pi_y\) is indistinguishable from the real distribution map in the regions that are important for constructing an optimal classifier, which will be discussed in Section~\ref{sssec:constructing-outcome-performativity}.
    Notably, POI-Boost is a non-parametric approach, based on stitching together copies of circuits,
    % and so does not incur a misspecification error.
    and so can achieve an arbitrarily low approximation error.
    However, as with any high-capacity model, one has to contend with the risk of overfitting.

\subsection{Challenges in dataset collection}%
\label{ssec:overview-of-some-common-datasets}

%\javi{Somehow, performative prediction makes datashift distributions after deployment intention-free. It just models the distribution shift mathematically (that is why is an umbrella). This is very interesting but sometimes far from practice. Each of the related fields uses one intention to somehow specify the problem and make it realistic. Maybe we have to justify why as soon as we specify the distribution map in practice, we go from performative prediction to other fields. Also, maybe the (ambitious) position that we want to take is to say that is impossible (difficult) to do practical work on performativity because as soon as you define the distribution map in a practical way, you are defining an intention and therefore, you fall into another field.}
%

As seen throughout this survey, the performative prediction literature has had a theoretical perspective. In Sections~\ref{sec:opt-perf-stable} and \ref{sec:opt-perf-opt}, we will see how the main focus of the literature is to give convergence guarantees. Therefore, experiments have mainly been used to complement the theoretical results. There is only a small part of the literature performing real world experiments. This situation is mainly due to the difficulties stated in the introduction of Section~\ref{sec:distribution-map}: once a model is deployed, it is locked in and one can \textit{only} observe the distribution shift caused by that model. This, added to the fact that the distribution shift happens slowly in practice once the model is deployed (think again about the bank example: individuals might take even years to modify their financial details before reapplying), makes it very difficult to perform real world experiments.

Outcome performativity is the only setup where these problems are not that restrictive (see Section~\ref{sssec:reuse-labels}). As the input features $x$, do not change due to the distribution shift, and, only the marginal distribution $p(y|x)$ changes. There are papers in this direction which can be seen as studies with real datasets~\cite{perdomo2025difficult,munro2025treatment} although performative prediction has not been their main focus of and their datasets are not publicly available.

Nevertheless, the main body of the literature of performative prediction is still far from this setup and has used semi-synthetic datasets to overcome the problems. These semi-synthetic datasets are created by equipping an existing dataset with a mechanism for reacting to the classifier, i.e., an initial dataset is combined with a (synthetic) distribution map\footnote{Some works have performed real-world experiments to obtain a better informed model of the distribution map~\cite{munro2025treatment}}.

Following the example by the seminal work by \citet{perdomo2020}, many of the works use strategic classification\footnote{See Section~\ref{sssec:strategic-classification} for details of this neighbouring field)} for this objective. 
See Appendix~\ref{appendix:commonly-used-datasets} for a presentation of commonly-used datasets for performative prediction.

%%% Local Variables:
%%% mode: LaTeX
%%% TeX-master: "../main"
%%% End:

% LocalWords:  eq Reparameterizable RGD RRM
          % Section 3: Datasets
\section{Optimizing for performative stability as a proxy}%
\label{sec:opt-perf-stable}
As discussed in Section~\ref{ssec:performative_optimality_and_stability},
the deployer is usually ultimately interested in minimizing the performative risk, $\mathit{PR}$,
which corresponds to finding the optimal point, $\theta_\mathit{PO}$.
However, as mentioned before,
finding the \emph{stable} point is easier and under certain conditions the stable and the optimal point coincide, as established by Theorem~\ref{thm:approximation} below.

In this section, we discuss the parts of the literature which are optimizing for performative stability
as a proxy for performative optimality.
We keep the discussion of the theorems proposed in prior work somewhat short
as our main focus in this survey is to give on overview of the many flavours of performative prediction approaches..
For a survey that presents the main convergence proofs in detail see~\cite{hardt2025performative},
written by co-founders of the field.

This section is framed as ``stability as an approximation of optimality''
but there are also reasons to prefer stability over optimality.
We already mentioned that stability is easier to certify (Section~\ref{ssec:performative_optimality_and_stability}) and the main optimization algorithms simply require retraining.

\subsection{Repeated risk minimization}%
\label{ssec:repeated-risk-minimisation}

\subsubsection{Prerequisites}%
\label{sssec:prerequisites}
Finding the stable point is easier because it can be found by an iterative procedure that only needs to sample from the new distributions and does not need to be aware of the inner workings of the distribution map.
In each iteration, we deploy new model weights, \(\theta\), which trigger a distribution shift. After collecting samples from the new distribution, we train a new model.

    As it can require many iterations to reach the stable point,
    and each iteration requires many samples from the current training distribution,
    this approach is only feasible if the samples can be obtain cheaply.

According to the levels of access as defined in Section~\ref{sssec:levels-of-access}, this requires at least Level 1b access (``cheap samples'').
Level 1a access, where getting samples is costly, is not practical for this scenario.

\subsubsection{Finding the stable point}%
\label{sssec:finding-stable}
Recall that the performatively stable points are defined as the fixed points of the function
\begin{equation}
    g(\theta)=\argmin_{\theta'\in\Theta} \;\mathrm{Risk}(\theta', \gD(\theta))~.
    \label{eq:fixed-point}
\end{equation}
This suggests the following procedure for finding the stable point.
Start with an arbitrary initialization of the model, \(\theta_0\in \Theta\), and let
\begin{equation}
    \theta_{t+1}:=\argmin_{\theta\in\Theta} \;\mathrm{Risk}(\theta, \gD(\theta_t))~.
    \label{eq:iteration}
\end{equation}
That is, for every step of the procedure,
we find the optimal model on the distribution induced by the previous model.
We call this procedure \emph{repeated risk minimization} (RRM).

Repeated risk minimization hopefully makes intuitive sense,
but the question remains whether it actually converges
and the next question after that will be whether the stable point is close to the optimal point.
The convergence of RRM is established by Theorem 3.5 in~\citet{perdomo2020}.
We replicate the statement of the theorem here,
but in the form it appeared in~\citet{hardt2025performative}.
The theorem makes mention of specific smoothness and sensitivity assumptions,
which are defined below.
Strong convexity refers to the usual definition.
\begin{theorem}
    If the loss function \(\ell(\theta, z)\) is \(\gamma\)-strongly convex
    and \(\beta\)-jointly smooth,
    then, repeated retraining defined in~\eqref{eq:iteration} converges to a unique stable point
    as long as the \(\epsilon\)-sensitivity of the distribution map \(\mathcal{D}(\cdot)\) satisfies \(\epsilon<\frac{\gamma}{\beta}\).
    Furthermore, the rate of convergence is linear, and for \(t\geq 1\), the iterates satisfy
    \begin{equation}
        \|\theta_t-\theta_\mathit{PS}\|\leq\left(\frac{\epsilon\beta}{\gamma}\right)^t \|\theta_0-\theta_\mathit{PS}\|~.
        \label{eq:theorem-iterates}
    \end{equation}
\end{theorem}
Throughout, \(\|\cdot\|\) denotes the \(L_2\)-norm.
The \(\epsilon\)-sensitivity of the distribution map is a statement about how much the distribution changes
in response to small changes in the the model weights:
\begin{definition}[Sensitivity]
    We say the distribution map \(\mathcal{D}(\cdot)\) is \(\epsilon\)-\emph{sensitive} if for all \(\theta, \theta'\in\Theta\),
    it holds that
    \begin{equation}
        \mathcal{W}(\mathcal{D}(\theta), \mathcal{D}(\theta'))\leq \epsilon\|\theta-\theta'\|~,
        \label{eq:sensitivity}
    \end{equation}
    where \(\mathcal{W}\) denotes the Wasserstein-1 distance.
The \(\epsilon\) parameter in \eqref{eq:sensitivity} has been dubbed the \emph{performative effect}~\cite{mendler2020stochastic}
because a large \(\epsilon\) allows large changes in the distribution in response to small changes in the model.%
    \label{def:sensitivity}
\end{definition}
The loss function, \(\ell\), needs to be smooth in the input \(z\),
but also smooth in the weights, \(\theta\),
We express this with \emph{joint smoothness} (defined in Appendix~\ref{appendix:definitions}).
If any of the three given assumptions (strong convexity, joint smoothness, and sensitivity) is not met,
repeated risk minimization has no guarantees for convergence.

Even when all assumptions are met, the procedure only converges in the infinite limit,
but in practice, we set a convergence threshold and stop once updates to the model weights are below the threshold.

\subsubsection{Approximation of optimality}%
\label{sssec:approximation-of-optimality}
We have seen the conditions under which repeated risk minimization converges to a performatively stable point.
The next question is how well this approximates the optimal point.
A bound for the difference between the two is established by Theorem 4.3 in~\citet{perdomo2020}.
\begin{theorem}%
\label{thm:approximation}
    If the loss \(\ell(z;\theta)\) is $L_z$-Lipschitz in z,
    \(\gamma\)-strongly convex,
    and the distribution map \(\mathcal{D}(\cdot)\) is \(\epsilon\)-sensitive,
    then, for every performatively stable point \(\theta_\mathit{PS}\)
    and every performative optimum \(\theta_\mathit{PO}\):
    \begin{equation}
        \|\theta_\mathit{PO}-\theta_\mathit{PS}\|\leq \frac{2L_z\epsilon}{\gamma}~.
        \label{eq:stable-optimal-gap}
    \end{equation}
\end{theorem}
As we can see, the quality of the approximation depends again on the smoothness and convexity
of the loss function and the distribution map.

\subsection{Repeated gradient descent}%
\label{ssec:repeated-gradient-descent}
In repeated risk minimization,
we perform the full optimization for the previous distribution, \(\mathcal{D}(\theta_{t-1})\), at every step.
That is, at each step, we find the \(\theta\) which minimizes \(\mathrm{Risk}(\theta, \mathcal{D}(\theta_{t-1}))\).
This might seem wasteful, as we have to start from scratch again at the next time step,
where the optimization problem is a different one.

Instead of fully optimizing \(\theta\) at every time step,
we can also perform a limited optimization.
This is where \emph{repeated gradient descent} (RGD) was proposed~\cite{perdomo2020}.
It performs just one gradient update for every time step \(t\).
Under certain conditions, RGD also converges linearly to a stable point.
See~\citet{perdomo2020} for details.

Note, however, that the number of evaluations needed of the distribution map, \(\mathcal{D}\), is not reduced in RGD,
so an efficient simulation of \(\mathcal{D}\) --- i.e., level 1b access (``cheap samples'') --- is still required.

\subsection{Stochastic Methods}%
\label{ssec:stochastic-methods}
The motivation for studying \emph{stochastic} optimization in performative prediction comes from the fact that the institution who is training the model cannot sample from the whole distribution instantaneously. Instead, they typically only observe a subset of samples before having the need for retraining—for example, a bank only observes how its lending rule impacts the population when applicants return to reapply.

\citet{mendler2020stochastic} formalize this by analysing learning dynamics where the distribution responds to the model one sample at a time. They propose two update schemes. In \emph{greedy deployment}, the model is updated and immediately redeployed after each observed data point, leading to a ``fully online'' feedback loop. In contrast, \emph{lazy deployment} performs the same incremental SGD updates per sample (one SGD update per sample) but only redeploys the updated model after accumulating $n(k)$ samples at iteration $k$. Under assumptions similar to those in \citet{perdomo2020}, they show that both schemes converge to a performatively stable point, although the convergence behaviour differs depending on how frequently redeployment occurs.

The work by \citet{drusvyatskiy2023stochastic} is more comprehensive as it considers the convergence to a stable point with a variety of classic stochastic optimization algorithms: such as stochastic gradient, clipped gradient, proximal point, and dual averaging methods, along with their accelerated and proximal variants.

\citet{li2024stochastic} further extend these guarantees to settings where the loss may be smooth but non-convex. This generalization is important because real-world learning problems rarely satisfy global convexity conditions. However, without convexity, one cannot expect convergence to an exact performatively stable point. Instead, their results show convergence to \emph{stationary performatively stable} solutions ($\theta_{\delta-SPS}$), points where the gradient of the performative loss is small (less than $\delta$) but not necessarily zero.
\begin{align}
    \|\nabla_\theta Risk(\theta_{\delta-SPS}, \mathcal{D}(\theta_{\delta-SPS}))\|^2 = \| \mathbb{E}_{z\sim\mathcal{D}(\theta_{\delta-SPS})}[\nabla l(z,\theta_{\delta-SPS})]\|^2 \le \delta
\end{align}
These solutions capture the idea of being “good enough” in non-convex settings, where the model and the induced distribution are approximately stable even if exact equilibrium cannot be reached.

\citet{wood2021online} proposes an online projected stochastic gradient descent algorithm for when the cost and the distribution map vary over time.

\citet{li2024clipped} studied the converge of the case when the gradient is clipped before applying the gradient descent step. This behaviour is important in several applications such as privacy preserving optimization, neural network training, etc. This algorithm also converges to a stable point for the standard conditions assumed in performative prediction. 

\subsection{Settings with multiple deployers}%
\label{ssec:settings-with-multiple-deployers}
    The original setup of performative prediction only considers one institution --- one deployer ---
    but several works have considered the setup with multiple deployers.
    In the bank loan example, we could for instance consider the case where there are multiple banks.
    In the works that we describe in the following,
    there are three different ways to construct the scenario with multiple deployers:
    (i) multiple deployers with one global distribution:
    in this case, the banks compete for one homogeneous population of customers;
    (ii) multiple deployers with globally-affected local distributions:
    in this case, each bank has its own local population of customers, but the customers also pay attention to what the other banks are doing;
    and (iii) multiple deployers with locally-affected local distributions:
    in this case, each bank has its own completely isolated customer population.
    The last scenario has an additional requirement in order to not make it simply be a collection of unrelated performative prediction problems:
    the deployers wish to all deploy the same model.
    In the banking example, this could be motivated by wanting to standardize their predictive models.
    % The setups presented in those works differ mainly in two axes~\cite{li2022multi}:
    % (i) the axis of whether there is one global distribution that is the same for every deployer
    % or whether deployers have their own local distribution;
    % and (ii) if there are local distributions, whether they are affected by any model deployment
    % or whether they are affected only by models of the local deployer.

\subsubsection{Multiple deployers with one global distribution}%
\label{sssec:mutli-one-global}
    \citet{piliouras2023multi} present a setup where there are \(n\) agents (deployers)
    with their own model parameters: \(\theta^1, \dots, \theta^n\).
    There is a single distribution map, \(\mathcal{D}\), but it depends on all the models:
    \(\mathcal{D}(\theta^1, \dots, \theta^n)\).
    We use the shorthand \(\bm{\theta} := (\theta^1, \dots, \theta^n)\).
    The risk for agent \(i\) is given by
    \begin{align}
        \mathrm{Risk}(\theta_i, \mathcal{D}(\bm{\theta})) =
        \mathbb{E}_{z\sim \mathcal{D}(\bm{\theta})}\big[\ell(z; \theta_i)\big]~.
        \label{eq:simple-multi-agent-risk}
    \end{align}
    %
    % An example that is described by this setup is a situation where, as in the example in Section~\ref{ssec:guiding-examples},
    % we have car navigation systems, but instead of everyone using the same system, there are multiple competing systems,
    % used by different drivers.
    % The traffic on the road is then affected by all of these systems.
    % The bank loan example can also be seen as a multi-agent setup, when we consider multiple banks.
    The authors define two different notions of performative stability for this multi-agent setting:
    The model vector \(\bm{\theta}\) is \emph{ex-ante performatively stable} if
    \begin{align}
        \theta^i = \argmin_{\theta_i\in \Theta}\mathrm{Risk}\big(\theta^i, \mathcal{D}(\bm{\theta})\big),
        \qquad\text{ for all } i\in\{1, \dots, n\}
    \end{align}
    where \(\theta_i\) is an element of \(\bm\theta\) and where \(\bm\theta\) is \emph{fixed}.
    This is a straight-forward generalization of performative stability to the multi-agent setting.
    Conversely, a model vector \(\bm\theta\) is \emph{ex-post performatively stable} if
    \begin{align}
        \theta^i = \argmin_{\theta_i\in \Theta}\mathrm{Risk}\Big(\theta_i, \mathcal{D}\big((\bm{\theta}^{- i}, \theta^i)\big)\Big),
        \qquad\text{ for all } i\in\{1, \dots, n\}
    \end{align}
    where \((\bm{\theta}^{- i}, \theta^i)\) is \(\bm\theta\) with the \(i\)-th model replaced by \(\theta^i\).
    This definition has more in common with the definition of the performatively optimal point,
    because we optimize over both the model used for making predictions and the model used to generate the data distribution.
    However, only one element of \(\bm\theta\) is optimized at a time,
    so it cannot be considered to be a definition of performative optimality.

    \citet{piliouras2023multi} define a multi-agent performative \emph{optimal} point simply as the point that minimizes the sum of the individual performative risks.
    %
    % \begin{align}
    %     \bm{\theta}^* = \argmin _{\bm\theta\in \Theta^n} \sum _{i=1}^n \mathrm{Risk}\big(\theta_i, \mathcal{D}(\bm\theta)\big)
    % \end{align}
    %
    Their analysis shows that under certain conditions
    (among other things, assuming a location-scale family for the distribution map),
    all three points (ex-ante and ex-post performatively stable and performatively optimal) coincide.
    The authors furthermore analyse when the multi-agent setup converges and when it chaotic dynamics.

\subsubsection{Multiple deployers with globally-affected local distributions}%
\label{sssec:mutli-local-globally-affected}
    \citet{narang2023multiplayer} consider a setup where there are \(n\) agents (deployers)
    and corresponding \(n\) local distributions, which, however, depend on the deployed models of \emph{all} agents.
    We again use the shorthand \(\bm{\theta} := (\theta^1, \dots, \theta^n)\).
    The risk for agent \(i\) is given by
    \begin{align}
        \mathrm{Risk}(\theta^i, \mathcal{D}_i(\bm\theta))=
        \mathbb{E} _{z\sim \mathcal{D}_i(\bm\theta)}\big[\ell(z; \theta^i)\big]~.
    \end{align}
    In contrast to \eqref{eq:simple-multi-agent-risk}, each agent has their own \(\mathcal{D}_i\) here.
    % An example of a multi-deployer setting where each deployer encounters a different distribution
    % is a set of universities where each university caters to different student populations.
    % They might, for examples, be universities specializing in liberal arts versus science and engineering;
    % and yet, students might apply across many universities,
    % such that they come into contact with many decision systems
    % and such that their applications are influenced by all of them.

%
    \citet{narang2023multiplayer} use the same stability concepts as \citet{piliouras2023multi} --- ex-ante and ex-post performative stability ---
    but refer to them as \emph{performatively stable equilibrium} and \emph{Nash equilibrium} respectively.
    % We can again define two different kinds of stability in this setting:
    % one in which the distribution is held fixed (corresponding to \emph{ex-ante} performative stability above),
    % which \citet{narang2023multiplayer} simply call \emph{performatively stable equilibria};
    % and one in which we consider the effect of one model at a time on the data distribution
    % (corresponding to \emph{ex-post} performative stability above),
    % which the authors refer to as the \emph{Nash equilibrium}.
    The name of the latter stems from the fact that in this solution point, agents have no incentive to change their deployed model,
    while taking the effect on the data distribution into account.

    The authors' analysis shows that performatively stable equilibria exist and are unique
    under the typical assumptions of smoothness and convexity.
    Repeated retraining (Section~\ref{ssec:repeated-risk-minimisation})
    and stochastic gradient methods (Section~\ref{ssec:stochastic-methods}) will converge to these equilibria.
    The authors additionally develop conditions and methods for reaching the Nash equilibrium.

    \citet{wang2023network} generalize this setup by considering a graph of deployers where only those deployers connected by an edge are influencing each other's local distribution.
    For the case of a fully-connected graph this is equivalent to \citet{narang2023multiplayer}'s setup.

\subsubsection{Multiple deployers with locally-affected local distributions}%
\label{sssec:mutli-local-locally-affected}
    \citet{li2022multi} consider a setup where \(n\) deployers each have their own distribution maps \(\mathcal{D}_i\),
    which are only affected by the local deployment: \(\mathcal{D}_i(\theta^i)\).
    The twist is, however, that these deployers want to come to a consensus such that everyone deploys the same model:
    \(\theta^i=\theta^j\), for all \(i\) and \(j\).
    What the authors are interested in in finding the \emph{multi-agent performatively stable} solution,
    in which they give all the distribution maps \(\mathcal{D}_i\) equal weight to define the following optimization problem:
    \begin{align}
        \theta_ \mathit{MPS} = \argmin _{\theta\in\Theta} \frac1n\sum _{i=1}^n\mathbb{E}_{z\sim \mathcal{D}_i(\theta_ \mathit{MPS})} \big[\ell(z; \theta)\big]
    \end{align}
    where \(\theta_ \mathit{MPS}\) is the multi-agent performatively stable point.

    The authors provide necessary and sufficient conditions for the existence and uniqueness of such a solution.
    Furthermore, they provide a practical algorithm based on \emph{decentralized stochastic gradient descent} (DSGD)
    a decentralized learning algorithm~\cite{lian2017can}.

\subsection{Convergence in other setups}
\subsubsection{Stateful distribution maps}%
\label{sssec:stateful-distribution-maps-methods}
As mentioned in Section~\ref{ssec:stateful-performative-prediction},
there are works that have explored setups in which the distribution map depends not only on the model parameters, $\theta$, but also on the distribution of the previous time step --- i.e., the distribution map depends on the previous state of the world.

We will briefly describe here under what conditions RRM converges in this setting
and when a limiting distribution exists, based on the analysis performed by \citet{brown2022performative}.

Instead of searching for a stable point, \(\theta_ \mathit{PS}\),
we are searching for a ``stable pair'': \((\mathcal{D}_ \mathit{SPS}, \theta_ \mathit{SPS})\),
as defined previously in Section~\ref{ssec:stateful-performative-prediction}.

    We can still apply RRM in stateful performative prediction:
    \begin{align}
        (Q_t, \theta_{t+1}) = \Big(\mathrm{Tr}(\theta_t, Q _{t-1}),\; \argmin\nolimits_ \theta \mathrm{Risk}\big(\theta_t, \mathrm{Tr}(\theta_t, Q _{t-1})\big)\Big)~.
    \end{align}
    In other words, the model parameters \(\theta _{t+1}\) are optimized on distribution \(Q_t\).
    To ensure that RRM converges to a stable pair,
    we again require that the loss function \(\ell(\theta, z)\) is \(\gamma\)-strongly convex
    and \(\beta\)-jointly smooth.
    Additionally, the transition map, \(\mathrm{Tr}\), needs to be \(\varepsilon\)-jointly sensitive,
    which is a generalization of \(\epsilon\) sensitivity as defined in Definition~\ref{def:sensitivity} above:
    A transition map \(\mathrm{Tr}(\cdot, \cdot)\) is \(\epsilon\)-jointly sensitive
    if for all \(\theta, \theta'\in \Theta\) and \(Q, Q'\in \Delta(\mathcal{Z})\)
    \begin{align}
        \mathcal{W}(\mathrm{Tr}(\theta, Q), \mathrm{Tr}(\theta', Q'))\leq
        \epsilon\mathcal{W}(Q, Q') + \epsilon\|\theta-\theta'\|_2
    \end{align}
    where \(\mathcal{W}(\cdot, \cdot)\) denotes the Wasserstein distance as before.
    Under these assumptions, RRM converges to a unique fixed point which is a stable point (Theorem 4 in \citet{brown2022performative}).

    We can also make a statement similar to that expressed in \eqref{eq:stable-optimal-gap},
    about how close the stable point will be to the optimal point, as defined in \eqref{eq:long-term-perf-optimal}.
    It requires that we additionally assume that \(\ell(z;\theta)\) is \(L_z\)-Lipschitz (Theorem 6 in \citet{brown2022performative}):
    \begin{align}
        \|\theta _{\mathit{SPO}} - \theta _{\mathit{SPS}}\| \leq
        \frac{2L_z \epsilon}{\gamma(1-\epsilon)}~.
    \end{align}

    \citet{li2022state} combine the stateful setup with stochastic methods (see Section~\ref{ssec:stochastic-methods} above).
    They follow the greedy deployment setup~\cite{mendler2020stochastic} where the model is deployed after each observed data point.
    The statefulness of the distribution map is then not modelled as a transition map on distribution over \(\mathcal{Z}\)
    but as a (\(\theta\)-dependent) Markov chain on individual samples: \(z _{k+1}\sim P_\theta(z_k)\).
    \citet{liu2024twotimescale} consider the same Markov-chain setting,
    but aim to find a (\(\epsilon\)-)\emph{stationary} point --- instead of a stable point ---
    which is defined as \(\|\nabla \mathcal{L}(\theta)\|^2 \leq \epsilon\) with derivative-free optimization.

\subsubsection{Model-output-dependent distribution maps}%
\label{sssec:convergence-model-output-dist-map}
As mentioned in Section~\ref{sssec:dependence-on-the-model-weights},
there are works in which it is assumed that the distribution map does not depend on the model parameters directly,
but rather on the observable model \emph{behaviour}:
given some inputs, what are the outputs?
For such a case, the theoretical analysis is different,
because we cannot talk anymore about the sensitivity of the distribution map \emph{w.r.t.} the model \emph{parameters}.
Rather, we have to consider the sensitivity \emph{w.r.t.} the model \emph{input-output behaviour}.

\citet{mofakhami2023performative} reformulate sensitivity of the distribution map and convexity and Lip\-schitz-ness of the loss function
in terms of model outputs instead of model parameters.
Let \(f_\theta\) refer to the function describing to the input-output behaviour of the model parameterized by \(\theta\);
i.e., if $z$ is the input to the model, then $f_\theta(z)$ is the output of the model.
Then the authors define \(\epsilon\)-sensitivity as (cf.\ \eqref{eq:sensitivity}):
\begin{equation}
    \chi^2\big(\gD(\theta), \gD({\theta'})\big)\leq \epsilon\mathbb{E}_{z\sim \mathcal{D}_\mathrm{base}}\big[\|f_\theta(z) - f_{\theta'}(z)\|^2\big]
    \label{eq:sensitivity-output-dependent}
\end{equation}
where Pearson's \(\chi^2\) divergence is used as the distribution distance function
instead of the Wasserstein function as before.
The reason for the change of the distance function is that with the Wasserstein distance,
it cannot be guaranteed that the repeated risk minimization will converge
(see Proposition 1 in~\citet{mofakhami2023performative}).

With this assumption and the other re-formulated assumptions (A1, A2, A3, and A4 in~\citet{mofakhami2023performative}),
RRM can be shown to converge even in the setting with model-output-dependent distribution maps.

The fact that the loss now needs to be convex \emph{w.r.t.} the model outputs instead of the model parameters,
makes convexity much easier to satisfy for neural networks,
which is one of the main motivations for this approach.

\subsubsection{Bilevel optimization}

Bilevel optimization \cite{colson2007overview} refers to solving one optimization problem while treating the solution to another, nested problem as a constraint. It has become very relevant in modern machine learning, as it often appears in settings like continual learning, where we want the final solution to serve as a strong starting point for future tasks.

\citet{pmlr-v202-lu23a} introduce \emph{bilevel performative prediction} by extending the notions of performatively stable and performatively optimal points to bilevel optimization settings. By adapting the assumptions of the original performative prediction framework~\cite{perdomo2020} to the bilevel case, they obtain convergence guarantees for bilevel RRM analogous to those established for the single-level setting. Moreover, their work extends the stochastic optimization results of \citet{mendler2020stochastic} to the bilevel regime, showing that similar stability and convergence behaviour can be achieved under stochastic updates.

%%% Local Variables:
%%% mode: LaTeX
%%% TeX-master: "../main"
%%% End:
       % Section 4: Performative stability
\section{Optimizing for performative optimality}%
\label{sec:opt-perf-opt}
While \emph{performative stability} is both easier to certify and to achieve,
what the model deployer is usually interested in is performative optimality ---
finding the model parameters \(\theta_ \mathit{PO}\) such that the performative loss \(PR(\theta_ \mathit{PO})\) is minimal.
Moreover, the assumptions that the optimal point and stable point lie in the same neighbourhood can be false even for simple distribution maps. In those cases, retraining approaches (that are used to get stable points) are provably sub-optimal~\cite{kabra2024limitations}.

Many approaches to reach the optimal point have been proposed in the literature,
but usually they do not compete with each other ---
rather, they are all aimed at different situations where different assumptions hold.

\begin{figure}
    \centering
    \begin{subfigure}[b]{0.72\textwidth}
        \includegraphics[width=\linewidth]{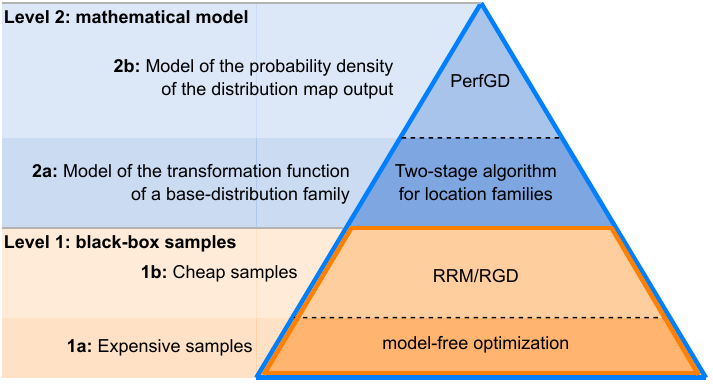}
        \caption{Diagram of the access pyramid.}
        \label{fig:access-pyramid-diagram}
    \end{subfigure}
    \begin{subfigure}[b]{0.24\textwidth}
        \centering
        \begin{tabular}{cl}
            \toprule
            \textbf{Level} & \textbf{References} \\
            \toprule
            2b             & \cite{izzo2021learn} \\
            \\
            \\
            \midrule
            2a             & \cite{miller2021outside,kim2023making,cyffers2024optimal} \\
            \\
            \\
            \midrule
            1b             & \cite{mendler2020stochastic,drusvyatskiy2023stochastic,li2024stochastic} \\
            &\cite{perdomo2020}\\
            \midrule
            1a             & \cite{jagadeesan2022regret,chen2024banditfeedback} \\
            \bottomrule
        \end{tabular}
        \caption{%
            Papers as examples for the levels.
        }%
        \label{tab:access-pyramid}
    \end{subfigure}
    \caption{%
        The \emph{access pyramid} establishes what methods can be applied depending on the level of access to the distribution map, as defined in Section~\ref{sssec:levels-of-access}.
      Methods that only require a lower level can also be applied when a higher level of access is available:
      if we, for example, have access to the model of the dynamics, then we can also produce samples.
      Working on level 2 usually requires fitting a mathematical model to collected samples, as described in Section~\ref{ssec:distribution-map-model-fitting}.
    }%
    \label{fig:access-pyramid}
    % Description of the figure for visually-impaired readers:
\end{figure}

The basic difficulty in optimizing for the performative loss $PR$ is
that we have to optimize over the distribution map \(\mathcal{D}(\cdot)\),
because we have to find the data distribution \(\mathcal{D}(\theta')\) which admits the lowest loss.
There are two fundamental paths to take here:
(i) make only general regularity assumptions about the distribution map;
(ii) assume access to a mathematical model of \(\mathcal{D}(\cdot)\) that we can directly use in the optimization.
    In practice, option (ii) requires fitting a mathematical model of the distribution map to observed samples
    as described in Section~\ref{ssec:distribution-map-model-fitting}.
    Fig.~\ref{fig:access-pyramid} gives an overview of which level of access (Section~\ref{sssec:levels-of-access}) is needed for which method.

\subsection{Performative optimality with regularity assumptions}%
\label{ssec:model-free}
\subsubsection{Bandits assuming sensitivity and smoothness}%
\label{sssec:bandits-assuming-sensitivity-and-smoothness}
To start with, we formulate performative prediction as a bandit setup.
At every time \(t\), the deploying institution chooses a model \(\theta_t\)
and observes a constant number of samples:
\begin{equation}
    \{z_t^{(i)}\}_{i\in[m_0]}, \text{ where }z_t^{(i)}\sim \mathcal{D}(\theta_t)~.
    \label{eq:get-samples}
\end{equation}
As is typical for bandit problems, we define a regret: \(\Delta(\theta_t)\coloneqq PR(\theta_t)-PR(\theta_ \mathit{PO})\).

\citet{jagadeesan2022regret} show that in such a setup,
and only assuming \(\epsilon\)-sensitivity of \(\mathcal{D}(\cdot)\) and Lipschitz-ness of \(\ell(z; \theta)\),
an algorithm can be constructed with a regret bound that outperforms the baseline Lipschitz bandit regret bound.

The basic idea of a continuum-armed bandit where the reward function is Lipschitz
is that we can exploit the fact that we have a bound on the rate of change,
to make guesses about the reward in the unexplored regions.
In our case, the reward function is the performative risk \(PR(\theta)\),
which is Lipschitz if \(\ell(z; \theta)\) is Lipschitz in both arguments
and \(\mathcal{D}(\cdot)\) is \(\epsilon\)-sensitive (Lemma 2 in~\cite{jagadeesan2022regret}).
However, in the case of performative prediction we can get a tighter bound.

The idea is that for a given distribution \(\mathcal{D}':=\mathcal{D}(\theta')\),
we can cheaply evaluate the risk for many models \(\theta\): \(\mathrm{Risk}(\theta, \mathcal{D}')\).
This is cheap because once we have deployed \(\theta'\) and have computed \(PR(\theta')\),
we necessarily must have all the information needed to compute \(\mathrm{Risk}(\theta, \mathcal{D}')\),
including the necessary data \(\{z_i\}\) and, if applicable, the necessary labels for computing the risk.

The key insight is then that we can get a bound for \(PR(\theta')\) by exploring other \(\theta\):
\begin{equation}
    PR(\theta') \leq \mathrm{Risk}\big(\theta, \mathcal{D}(\theta')\big) + L_z\epsilon\|\theta-\theta'\|
    \label{eq:bandit-bound}
\end{equation}
assuming \(L_z\)-Lipschitz-ness of \(\ell\) in the first argument and \(\epsilon\)-sensitivity of \(\mathcal{D}(\cdot)\).
%
%See also Fig.~\ref{fig:bandit}.

%\begin{figure}
%    \begin{center}
%        \includegraphics[width=0.9\textwidth]{figures/2025-02-27T22:12:32,053017926+01:00.png}
%    \end{center}
%    \caption{
%        \thomas{This is a screenshot of Figure 1 from~\cite{jagadeesan2022regret}; we need to recreate it.}
%        Illustration of the idea behind the bandit algorithm.
%    }%
%    \label{fig:bandit}
%\end{figure}

The usefulness of this bound is determined by which models \(\theta\) are explored.
We refer to~\cite{jagadeesan2022regret} for the detailed algorithm for optimal exploration.

\citet{chen2024banditfeedback} extend the bandit feedback setup to minimize the \textit{total} regret of deploying several models:
\begin{equation}
    \mathcal{R}_{N}(\mathcal{A}, PR) = \sum_{\tau=1}^{T_\mathit{total}}\sum_{i=1}^{n_{\tau}} l(z_{\tau}^{(i)};\theta_\tau) - N\cdot PR(\theta_{PO})
\end{equation}
So far, the assumptions made are very light, and exclude things like convexity of the loss.
This is a strength of the bandit algorithm,
however, with more assumptions, in particular with the \emph{location families} settings,
which we will discuss in more detail below, the bound can be strengthened further.

\subsubsection{Geometrically decaying environments}%
\label{sssec:geometrically-decaying-environments}
    Geometrically decaying environments have previously been defined in Section~\ref{sssec:transition-maps-from-distribution-maps}
    as a possible form of a transition map in \emph{stateful} performative prediction.
    They were investigated by \citet{ray2022decision}.
    Let \(Q_t\in\Delta(\mathcal{Z})\) be the distribution observed at a time \(t\).
    The transition map is given by \eqref{eq:geometric-decay}.

With additional assumptions on top of the usual assumptions of smoothness, convexity and sensitivity,
\citet{ray2022decision} present two algorithms for solving this setup.
One of the algorithms assumes the pure bandit setup where we only get ``zero-order'' information
(samples from \(Q_t\) and the corresponding risk).
The other assumes access to a gradient oracle which provides ``first-order'' feedback
in the form of \(\nabla_{z,\theta}\ell(z; \theta)\),
which is readily available if the model is differentiable.

\subsection{Performative optimality with explicit models}%
\label{ssec:model-based}
In Section~\ref{sec:opt-perf-stable},
we discussed methods which iteratively approach the stable point of the distribution map,
which under certain assumptions is close to the optimal point.
In Section~\ref{ssec:model-free},
we encountered methods that try to be more sophisticated in exploring the model space,
instead of just applying risk minimization repeatedly (as was the case in Section~\ref{sec:opt-perf-stable}).

However, ideally, we would use a more powerful optimization method, like gradient descent,
to find the optimal point of the distribution map.
However, for gradient descent, we require a differentiable model of the distribution map.
In other words, we need Level 2 access (``model of the distribution map''), as defined in Section~\ref{sssec:levels-of-access}.

Now, this does not mean that our differentiable model of the distribution map has to be written entirely by hand.
We can also define a differentiable hypothesis class with free parameters
and fit these parameters to samples we observe from the distribution map.
Then, once we have learned the model of the distribution map,
we can compute gradients through it, in order to find the performatively optimal model.
The point is that we have a model that allows the application of more efficient optimization methods.

% involves first constructing an explicit model of the distribution map, typically subject to some fairly strict assumptions (e.g. its belonging to the location family of distributions) and then optimising for performative optimality using said model

\subsubsection{Optimizing a model of the performative mechanism}%
\label{sssec:convex-optimization-with-location-scale-families}
We begin with~\citet{miller2021outside}, who aim to use \emph{convex optimization} to find the performatively optimal point.
In this work, the hypothesis class for the distribution map is \emph{location-scale families},
which we previously described in Section~\ref{sssec:location-scale-families}.

A requirement for using convex optimization is that the loss is convex,
and as we want the optimum with respect to the performative risk, $PR(\cdot)$,
we have to ensure that the performative risk --- which includes computing an expectation over the distribution map ---
as a whole is convex.
The authors identify \emph{mixture dominance} (equation A4 in~\cite{miller2021outside}) as the crucial property
that the distribution map has to satisfy to make the performative risk convex:
\begin{definition}[Mixture dominance]%
\label{def:mixture-dominance}
The pair \(\big(\mathcal{D}(\cdot),\ell(\cdot;\cdot)\big)\) of a distribution map and a loss function
satisfy mixture dominance if for all \(\theta, \theta', \theta_0\in\Theta\) and \(\alpha\in(0,1)\) we have:
    \begin{equation}
    \mathrm{Risk}\big(\theta_0; \mathcal{D}(\alpha\theta + (1-\alpha)\theta')\big)\leqslant
    \mathrm{Risk}\big(\theta_0; \alpha\mathcal{D}(\theta) + (1-\alpha)\mathcal{D}(\theta')\big)~.
    \end{equation}
\end{definition}
That is, the mixture distribution \(\alpha\mathcal{D}(\theta) + (1-\alpha)\mathcal{D}(\theta')\)
\emph{dominates} the distribution induced by a mixture model \(\alpha\theta + (1-\alpha)\theta'\),
under the loss function \(\ell\).

Together with previous smoothness assumptions
and the assumptions that \(\ell\) is (strongly) convex in both \(z\) and \(\theta\),
we get a convex performative risk.
Notably, distribution maps from location-scale families satisfy mixture dominance.
Location-scale families are not the only distribution families satisfying mixture dominance,
but they are the focus of the paper.

The authors propose picking a distribution map hypothesis class within the location-scale family and then fitting this to sampled data, as previously described in Section~\ref{ssec:distribution-map-model-fitting}.
In the most general case, we fit all parameters, \(\Sigma_0, \Sigma, \mu_0, \mu\), from \eqref{eq:location-scale-theta} to the observed data.
However, in the examples in the paper, only $\mu_0$ is assumed to be unknown and fit to observed samples.

With the estimated distribution map parameters, we can compute the empirical performative risk,
given samples from the base dataset \(\{z_i\}_{i=1}^N\) drawn from \(\mathcal{D}_ \mathrm{base}\):
\begin{equation}
    \widehat{PR}(\theta) = \frac1N\sum_{i=1}^N \ell\Big(\big(\hat{\Sigma}_0 + \hat{\Sigma}(\theta)\big)z_i +\hat{\mu}_0 + \hat{\mu}(\theta); \theta\Big)~.
    \label{eq:empirical-risk-location-scale}
\end{equation}
Here we replaced the expectation from~\eqref{eq:perform-risk} with the empirical mean,
and substituted \(z\) with the expression from~\eqref{eq:location-scale-theta}.
The expression in \eqref{eq:empirical-risk-location-scale} can be efficiently optimized with convex optimization
because it does not contain an expectation over \(\mathcal{D}(\theta)\), which contains a parameter we want to optimize.
Rather, the expression only contains an expectation over the base distribution \(\mathcal{D}_ \mathrm{base}\), which is fixed and not being optimized.
Therefore, this approach only needs a level 2a model (``model of the transformation function'') as defined in Section~\ref{sssec:levels-of-access}.
% As \(\Sigma\) and \(\mu\) (and thus also \(\hat{\Sigma}\) and \(\hat{\mu}\)) are linear maps, we can compute the gradient
% and use efficient gradient optimization methods to find the performative optimum.

The experiments by \citet{miller2021outside} show that such an approach converges much faster
than methods such as repeated gradient descent.
A caveat is that we assume that we can model the distribution map correctly.
    This approach was later generalized by \citet{cyffers2024optimal}
    who use the reparameterization trick to compute the performative risk gradient for any distribution from the base-distribution family.

\subsubsection{Fully differentiable model of the distribution map}%
\label{sssec:generic-differentiable-model-of-the-distribution-map}
A different approach is given by~\citet{izzo2021learn} (contemporary with \citet{miller2021outside}), which the authors call PerfGD.
Instead of considering distribution maps from a location-scale family,
they consider arbitrary differentiable distribution maps.
The method requires a 2b model (``model of the probability density'') as defined in Section~\ref{sssec:levels-of-access}).
Of course, such a model is usually not known \emph{a priori}, so \citet{izzo2021learn} assume that one has a hypothesis class of differentiable models of the distribution map, with free parameters that have to be learned from observed samples.

In the location-scale family setup above, we had parameters in the definition of the distribution map
that depended on \(\theta\): \(\Sigma(\theta)\) and \(\mu(\theta)\).
\citet{izzo2021learn} also have such parameters, but their exact form is not restricted.
Let \(f(\theta)\) be the parameters in our model of the distribution map that depend on \(\theta\).
We can then express the probability density of \(\mathcal{D}(\theta)\) over \(\mathcal{Z}\) as \(p(z;f(\theta))\).
We then require only that \(p(z;w)\) has a known functional form
and that it is differentiable in its second argument.
From \(f(\theta)\) we require that it is easily estimatable from samples drawn from \(\mathcal{D}(\theta)\)
for a fixed \(\theta\).
An example for the probability density is a mixture of Gaussians:
\begin{equation}
    p\big(z;f(\theta)\big) = \sum_{i=1}^K \gamma_i \mathcal{N}\big(z; f_i(\theta), \Sigma_i\big), \quad\text{where }\sum_{i=1}^K \gamma_i = 1
    \label{eq:mixture-of-gaussians}
\end{equation}
Here, \(f(\theta)\) returns a vector of the means of the Gaussians, while the covariances are fixed.
The mixture of Gaussians is differentiable in \(f\),
and \(f\) can be estimated efficiently from observed data, as required.

With this differentiable distribution map, we want to compute the gradient, with respect to \(\theta\), of the performative risk:
\begin{equation}
    \nabla_\theta PR(\theta)=\nabla_\theta \mathbb{E}_{z\sim \mathcal{D}(\theta)}\big[\ell(z; \theta)\big]
    \label{eq:pr-gradient}
\end{equation}
The challenge is handling the dependency on \(\theta\) in the distribution we are sampling from.
But with the assumption of differentiability of \(p(z; f(\theta))\) this is possible:
\begin{multline}
    \nabla_\theta PR(\theta)= \nabla_\theta\int_ \mathcal{Z} \ell(z; \theta) \,p\big(z; f(\theta)\big) \,\mathrm{d}z\\
    % = \int_ \mathcal{Z} \nabla_\theta \Big(\ell(z; \theta) \,p\big(z; f(\theta)\big)\Big) \,\mathrm{d}z\nonumber\\
    = \int_ \mathcal{Z} \frac{\partial\ell(z; \theta)}{\partial \theta} \,p\big(z; f(\theta)\big) \,\mathrm{d}z
    + \int_ \mathcal{Z} \ell(z; \theta)\,\frac{\partial f(\theta)}{\partial \theta} \left.\frac{\partial p(z; w)}{\partial w}\right|_{w=f(\theta)}\mathrm{d}z
    \label{eq:pr-gradient-derivation}
\end{multline}
The first integral in the final expression corresponds to the gradient
used in Repeated Gradient Descent (RGD) in Section~\ref{ssec:repeated-gradient-descent}.
We can turn the second integral back into an expectation over \(\mathcal{D}(\theta)\) with the log-derivative trick
(equation (3) in~\cite{izzo2021learn}):
\begin{equation}
    \nabla_\theta PR(\theta) = \mathbb{E}_{z\sim \mathcal{D}(\theta)}\left[\frac{\partial\ell(z; \theta)}{\partial \theta}
    + \ell(z; \theta)\,\frac{\partial f(\theta)}{\partial \theta} \left.\frac{\partial \log p(z; w)}{\partial w}\right|_{w=f(\theta)}\right]
    \label{eq:pr-gradient-final-step}
\end{equation}
For the details on how to estimate the various terms in practice, see the paper~\cite{izzo2021learn}.

The gradient in~\eqref{eq:pr-gradient-final-step} is referred to as the performative gradient in the paper.
The point of this gradient is that it can be used to optimize the performative risk itself and thus allows us to find the performatively optimal point.
RGD, in contrast, only uses the first term of the gradient for optimization --- the term corresponding to an ordinary supervised loss --- which can only find the \emph{stable point}.

    In a follow-up work, \citet{izzo2022learn} apply the idea of using the performative gradient
    to the problem of \emph{stateful performative prediction} as defined in Section~\ref{ssec:stateful-performative-prediction}.
    The goal is to minimize the long-term performative risk, \(\mathit{PR}_\infty\), (\eqref{eq:long-term-perf-risk}),
    which is given by the limiting distribution map, \(\mathcal{D}_\infty\), (\eqref{eq:limiting-distribution-map}), which is assumed to exist.
    It is further assumed that we have a parameterized model for the distributions \(Q\in \Delta(\mathcal{Z})\),
    such that \(p(z; \mu)\) is the distribution where \(\mu\) is a vector of parameters.
    This allows us to express the transition map, \(\mathrm{Tr}(\theta, Q)\),
    as a map on the parameters: \(\mu_t = m(\theta, \mu _{t-1})\),
    where \(\mu_t\) parameterizes the distribution at time \(t\).
    \citet{izzo2022learn} assume that \(\mu\) can be estimated from samples, but that the parameter map, \(m\),
    which represents the transition map, is unknown.
    This leads to the challenge that we need to know \(\mathcal{D}_\infty\) in order to compute the long-term performative risk,
    but \(\mathcal{D}_\infty\) in turn is defined by infinite applications of \(m(\theta, \mu _{t-1})\), which is unknown.
    \citet{izzo2022learn} show, however, that \(m(\cdot, \cdot)\) itself is not required to compute
    the \emph{long-term performative gradient}.
    Rather, only the partial derivatives with respect to the first and second argument are needed: \(\partial_1 m\) and \(\partial_2 m\).
    The paper presents an algorithm to estimate those derivatives from many observed triples \((\mu _{t-1}, \theta, \mu_t)\).

    The authors prove theoretical guarantees on how accurately the long-term performative gradient can be estimated,
    dependent on the number of deployments, where each deployment allows observing the transition map.
    As a deployment is expensive in a real-world context,
    it is desirable to use the information from each deployment as efficiently as possible.

\subsubsection{Constructing the optimal model from the distribution map model}%
\label{sssec:constructing-outcome-performativity}

    In the special case of \emph{outcome performativity} (see Section~\ref{sssec:outcome-perf-def}),
    we can explicitly construct the performatively optimal model directly from the learned model of the distribution map.
    Recall that in outcome performativity, we can learn a function \(q(x)\) in a format suitable for optimization which,
    together with the base distribution \(\mathcal{D}_ \mathrm{base}\),
    defines the distribution map.
    The function \(q(\cdot)\) represents \(P(y|\hat y, x)\), from which we can sample \(y\) conditioned on \(x\) and \(\hat y\).
    % From \(q\), we can use \eqref{eq:outcome-perf-q-pi} to construct a function \(\pi_y(x, \hat y)\),
    % which returns the conditional probability distribution \(P(y|\hat y, x)\).

%
    Where in the case of normal PP, the loss function has the form \(\ell(z; \theta)\) (where usually \(z=(x,y)\)),
    it now has the form \(\ell(x, y, \hat y)\),
    because in outcome performativity, the loss does not depend on the exact model parameters \(\theta\),
    and rather only on the predictions of our model, \(\hat y\).
    (For more discussion on this point, see Section~\ref{sssec:dependence-on-the-model-weights}).
    \citet{kim2023making} show then that the following definition of \(f^*\) defines a performatively optimal model:
    \begin{align}
        \forall x\in \mathcal{X},\quad f^*(x) = \argmin_{\hat y\in \mathcal{Y}} \mathbb{E}_{y\sim P(\cdot|\hat y, x)}\big[\ell(x, y, \hat y)\big]~.
        \label{eq:outcome-perf-opt}
    \end{align}
    There are two aspects to this claim:
    (i) the \emph{argmin} and expectation value in the equation are feasible to compute in practice;
    and (ii) \(f^*\) is indeed performatively optimal.
    Regarding (i): both the \emph{argmin} and the expectation value are only over 2 values in the binary case: \(\{0, 1\}\),
    so for a given \(x\in \mathcal{X}\), the loss function only has to be evaluated 4 times,
    which is practically feasible.
    For claim (ii), we write the performative risk, \(\mathit{PR}\), in terms of a model \(f\) instead of in terms of the model parameters, \(\theta\)
    (again, due to the fact that the distribution map in outcome performativity depends on model output, not model parameters):
    \begin{align}
        \mathit{PR}(f)=\mathbb{E}_{(x,y)\sim \mathcal{D}(f)}\big[\ell(x, y, f(x))\big]
        =\mathbb{E}_{x\sim \mathcal{D}_ \mathrm{base}}\Big\lbrack\mathbb{E}_{y\sim P(\cdot|\hat y=f(x), x)}\big[\ell(x, y, f(x))\big]\Big\rbrack~.
    \end{align}
    The second equality is valid because of the definition of the distribution map in outcome performativity
    as seen in \eqref{eq:outcome-perf-pi-dist-map}.
    We can see now that \(f^*\) in \eqref{eq:outcome-perf-opt} is defined such
    that it minimizes the contents of the outer expectation value on the left-most side.
    For a formal proof see \citet{kim2023making}.

%
    % Note that once the distribution map model \(q(\cdot)\) is obtained, no further errors are incurred in obtaining the performatively optimal solution.
    \citet{perdomo2025revisiting} has furthermore shown that outcome performativity allows finding a performatively stable solution under very mild assumptions (Theorem 4.1 in \citet{perdomo2025revisiting}),
    which do not include smoothness or \(\epsilon\)-sensitivity.

    An important advantage of this approach is that the loss function can be freely switched out \emph{after} training.
    Once \(\pi_y\) has been learned, an optimal classifier can be constructed for any loss \(\ell\).
    However, this requires that the learned distribution map model \(q\) is a sufficiently good model
    of the true distribution map for arbitrary regions of the \(\mathcal{X}\times \mathcal{Y}\) space,
    which is not necessarily the case and depends on how it was trained.
    The notion of constructing a classifier for any loss function after training
    was introduced as the concept of \emph{omnipredictors} by earlier work~\cite{gopalan2022omnipredictors} for the case of (non-performative) supervised learning.
    In practice, a space \(\mathcal{L}\) of loss functions can be decided upon before training,
    such that \(q\) is ensured to be a good distribution map model on that space \(\mathcal{L}\).
    For more details see \citet{kim2023making}.

%%% Local Variables:
%%% mode: LaTeX
%%% TeX-master: "../main"
%%% End:
      % Section 5: Performative optimality
\section{Cross-pollination}%
\label{sec:related-research-areas}
So far, we have dissected \textit{performative prediction} by systematically analysing its components: the distribution map and the optimization of the models.
During this process, we have seen how strategic classification has been useful for constructing concrete distribution maps for performative prediction.
However, there is plenty of more machinery devised in other fields that performative prediction, as an emerging field, can benefit from.
The hope is that there are opportunities for cross-pollination.
The aim of this section is to describe connections between more established fields and \textit{performative prediction}:
pointing out existing connections that have not been made explicit so far
% revise what what methods have been already used in \textit{performative prediction} without explicitly making that connection,
and making new connections in order to explore how performative prediction can be inspired by techniques in other fields in future research.
%These connections do not only include distribution map mechanisms, they also cover optimization algorithms or even concepts. 

%Connecting concepts (performative stability -> adversarial robustness). 

\begin{comment}
\begin{table}[tp]
    \centering
    \caption{%
    Comparison of different mechanisms for generating datasets for performative prediction.
    }%
    \label{tab:mechanism-comparison}
    \begin{tabular}{lccc}
         \toprule
                                               & Strategic      & Algorithmic & Adversarial \\
                                               & classification & recourse    & robustness  \\
         \midrule
         Finds an input which changes the output           & \cmark{} & \cmark{} & \cmark{} \\
         Input is restricted to be close to original input & \cmark{} & \cmark{} & \cmark{} \\
         Possibility for individualized cost functions     & (\cmark) & \cmark{} & \xmark{} \\
         Input is restricted to be realistic (non-OOD)     & \xmark{} & \cmark{} & (\cmark) \\
         Assumes that the model has found the real causal relation & & & \\
         between input and output                          & \xmark{} & \cmark{} & \xmark{} \\
         \bottomrule
    \end{tabular}
\end{table}
\end{comment}

\subsection{Alternatives to strategic classification}
In this section, we explore alternatives to strategic classification for generating performative-prediction datasets. These alternatives are meant to be used as \emph{drop-in replacements} for strategic classification, in the sense that they also modify individual samples, dependent on the deployed model.
On the other hand, mechanisms such as resampled-if-rejected (Section~\ref{sssec:resampled-if-rejected-procedure}) or outcome performativity (Section~\ref{sssec:outcome-perf-def}) are out-of-scope for this section, since they work very differently than strategic classification.
% We therefore exclude here the discussion of mechanisms that work very differently from strategic classification, like resampled-if-rejected (Section~\ref{sssec:resampled-if-rejected-procedure}) or outcome performativity (Section~\ref{sssec:outcome-perf-def}).

\subsubsection{Adversarial Attacks}

Adversarial attacks~\cite{szegedy2013intriguing} generate perturbed inputs that intentionally cause a classifier to misclassify. These \emph{adversarial examples} are defined with respect to a particular model $h_\theta$ and therefore depend directly on its parameters $\theta$. As a result, the process of crafting adversarial examples naturally induces a data distribution that is \emph{determined by the deployed model}. This is precisely the mechanism at the core of performative prediction.

Formally, constructing an adversarial example seeks the smallest perturbation $\delta$ that changes the model's prediction:
\begin{equation}
  x = x_{\mathrm{base}} + \delta \qquad \text{where} \qquad
  \delta = \argmin_{\delta} \|\delta\| \;\; \text{s.t.} \;\; h_{\theta}(x_{\mathrm{base}}+\delta) \neq h_{\theta}(x_{\mathrm{base}}).
  \label{eq:adversarial-robustness}
\end{equation}
This optimization implicitly defines a \emph{distribution map} $\mathcal{D}(\theta)$: the data observed after deployment is transformed in response to the model. Whether the adversary has access to model internals (white-box) or only input-output behaviour (black-box) mirrors the distinction in performative prediction between distribution maps that depend on the model parameters or only on its external predictions (see Section~\ref{sssec:dependence-on-the-model-weights}). We refer to \citet{chakraborty2018adversarial} for a survey of adversarial attacks.

The connection between adversarial attacks and performative prediction has recently begun to be explored empirically. For example, \citet{dong2023approximate} use adversarial responses as the distribution map in their experiments:
\begin{equation}
    x = x_{\mathrm{base}} - \varepsilon \nabla_x \mathrm{Risk}(x_{\mathrm{base}}; \theta).
\end{equation}
This corresponds to a basic gradient-based attack, whereas the adversarial robustness literature offers far more refined methods such as FGSM~\cite{goodfellow2014explaining} and DeepFool~\cite{moosavi2016deepfool}. We believe that these attacks could serve as more realistic and expressive distribution maps, improving the practicality of the field.
%, analogous to how strategic classification examples have been widely used in performative prediction experiments.

Beyond using adversarial examples to model distribution shifts, adversarial \emph{defences} also connect directly to optimization methods in performative prediction. The classic defence is adversarial training, which replaces training samples by their adversarial counterparts in gradient descent~\cite{madry2017towards} is equivalent to Repeated Gradient Descent in performative prediction (see Section~\ref{ssec:repeated-gradient-descent}), where the model is repeatedly updated to account for the distribution shift it induces.

This suggests a deeper conceptual correspondence: \emph{adversarial robustness}---the ability of a model to remain accurate under adversarial perturbations---parallels \emph{performative stability}.

% \footnote{We refer the reader to \citet{zhao2024adversarial} for a comprehensive survey}

%In fact, what the performative prediction literature calls Repeated Gradient Descent is adversarial robustness training, where in each step, the gradient is calculated using adversarial examples \thomas{citation?}.

%\thomas{If strategic classification and adversarial attacks are so similar, why not use adversarial robustness methods for addressing strategic classification?}

%\thomas{It seems one difference between performative prediction and adversarial robustness is actually that performative prediction only tries to defend against one specific mechanism that the users will use, whereas adversarial robustness tries to be generally robust.}

\subsubsection{Dataset generation based on algorithmic recourse}

Algorithmic recourse\footnote{Consult \citet{karimi2021algorithmic} for a survey} refers to the recommendations that are given to a ``data point'' to change the output of a specific model\footnote{Specialized literature uses the terms counterfactuals and algorithmic recourse interchangeable. We will follow this convention.}. What differentiates this field from previously discussed fields is that in general, in algorithmic recourse, the features of the data are not directly changed, they have to be changed through \textit{actions}. An \textit{action}, performed by an user to change their features, \emph{might} have an impact in more than one feature. This is intrinsically related to causality as several features might be causally related. 

Given a sample $x_\mathrm{base}$ and an action $a: \mathcal{X} \rightarrow \mathcal{X}$ that transforms a sample into another sample, an algorithmic recourse sample can be found with 
\begin{equation}
  x = a\big(x_\mathrm{base}\big)\quad \text{where} \quad a = \argmin_{a' \in \mathcal{A}}\; \operatorname{cost}\bigl(a'; x_\mathrm{base}\bigr) \quad \text{such that}\; h\big(a\big(x_\mathrm{base}\big)\big) \neq h\big(x_\mathrm{base}\big)
  \label{eq:algorithm-recourse}
\end{equation}
where ``$\operatorname{cost}(a; x_\mathrm{base})$'' is the cost of applying the action $a$ to $x_\mathrm{base}$ and $h(x)$ the output of the model for the input $x$.

The concept of the action is often associated with the requirement that the new input should be ``realistic'' which we can take to mean that the new input should lie within (or within some narrow radius of) the training distribution, which includes respecting the immutability or monotonicity (age, for instance, can only increase) of features.

% This stands in contrast to strategic classification and adversarial attacks, where the objective is to manipulate the classifier while leaving the underlying reality unchanged. For this reason, we see algorithmic recourse as a particularly promising source of inspiration for practical performative prediction experiments.
%
Apart from being realistic, these recommendations %
are also meant to preserve $P(y|x)$, the conditional relationship between the features, $x$, and the (true) outcome, $y$.
In other words, a change in $x$ may lead to a change in the true $y$.
 For example, in the bank loan example, the recourse can state that to receive a loan, a person should earn more money. If they changed jobs and started earning more, the person would, in fact, have more possibilities of repaying the loan due to their new financial stability.
If the conditional relationship of $x$ and $y$ were preserved, no detrimental effect on classifier performance would be observed.
However, \citet{konig2025performative} have explored cases where the post-recourse conditional distribution is changed. Previously, \citet{bracale2024learning} had used the concept of actions in performative prediction but not algorithmic recourse itself.

%\st{The procedure that users are following in strategic classification mirrors closely the process that is performed to generate recommendations in \emph{Algorithmic recourse} (AR), especially in the case where the utility function, $u$, captures whether the classification was positive or negative.}
%
%\st{In both AR and strategic classification (with binary classification target), we try to find an input to the classifier that is close to a base input $x_\mathrm{base}$ but which results in a different classification output (citation needed).}

%
%\st{A bad actor aiming to game the system would have little reason to respect for this constraint.}

\subsection{Overlaps with other fields}%
\label{ssec:miscellaneous-connections}

\subsubsection{Delayed impact as performative prediction}%
\label{sssec:delayed-impact-as-performative-prediction}
    The delayed impact framework by \citet{liu2018delayed}
    which is meant to show that interventions that were meant to help disadvantaged group (i.e., positive discrimination)
    can have long-term negative effects on those groups.
    The main example of the paper is a model of how the credit score (which is a feature) evolves
    in response to different classifiers.
    This model can be straight-forwardly re-formulated as a performative prediction setup
    (see Appendix~\ref{appendix:delayed-impact} for the details)
    and could thus be used as a dataset for the field of PP.

\subsubsection{Fairness under performative prediction}
Another question is how existing algorithmic fairness methods behave under performative prediction.
This question has received some attention in the literature.
\citet{mishler2022fair} find that even though a model can be trained to be fair, it can become unfair after the performative prediction setting, where there is a distribution shift.
\citet{jin2024addressing} reach a similar conclusion and propose a new algorithm to preserve the fairness throughout model deployments.
\citet{zezulka2023performativity} study under what conditions the model will stay fair under the performative prediction setting. 

\subsubsection{Performative reinforcement learning}%
\citet{mandal2023performative} consider a setup where the employed policy affects the reward and transition dynamics of the reinforcement learning environment.
While \citet{mandal2023performative} focuses on the tabular Markov decision process (MDP) setting,
their follow-up work~\cite{mandal2025performative} extends the analysis to the linear MDP setting.
Other work in this area extended the notion of performative reinforcement learning to other scenarios~\cite{pollatos2025corruption,rank2024performative}.

\subsubsection{Targeting calibration}%
\citet{oesterheld2023incentivizing} consider a special case of performative prediction where the model predicts the probability distribution of the environment --- i.e., the model tries to predict the output of the distribution map. This scenario describes an oracle-like AI\@.
It is relatively easy to show that most such distribution maps have fixed points, allowing the AI to make perfect predictions, but the score of such a prediction (which is equivalent to the negative performative risk) can still be arbitrarily bad.
The goal of this work is to find a suitable loss function for finding high-scoring fixed points.

    A very similar point was made by \citet{perdomo2025revisiting}.
    This paper is considering outcome performativity while being interested in achieving \emph{outcome indistinguishability} (or, equivalently, multicalibration~\cite{dwork2025fairness}), which means that the model output is indistinguishable from the environments distribution.
    Just like \citet{oesterheld2023incentivizing},
    the author finds that it is possible to make perfectly calibrated predictions, but that this in itself does not guarantee a low performative risk.

\subsubsection{Algorithmic collective action}
\citet{hardt2023algorithmic} investigated ways in which a group of users can affect what an institution's model learns.
By collectively deciding to change their public data in certain ways, they can --- among other things --- nudge the institution's model towards paying more or less attention to certain features.
The fact that the users' responses shape the final model is reminiscent of performative prediction.

%%% Local Variables:
%%% mode: LaTeX
%%% TeX-master: "../main"
%%% End:
       % Section 6: Related work
\section{Discussion and future directions}%
\label{sec:discussion-and-future-directions}
% \begin{itemize}
%     \item we have given an overview of the field
%     \item we laid out the different mechanisms and solutions, and categorised them
%     \item we have mention several avenues for transfers of ideas between different fields
%     \item the field is a bit strange because it's intentionally a very vague setup; when we consider more concrete problems, there are often more narrow methods that are better suited
%     \item we recommend the field define more strictly which kinds of scenarios it aims to address
%         \begin{itemize}
%             \item the car navigation example is probably better solved by a stochastic strategy
%             \item you probably shouldn't meddle in election forecasts: ``%
% For example, when building a model for election forecasting,
% such a model benefits from openness, simplicity and the ability to be audited,
% but such desiderata run counter to algorithms that anticipate voter shifts after publication of the forecast.''
%         \end{itemize}
%     \item researchers should try to find out what actually happens when these methods are applied
%         \begin{itemize}
%             \item one would hope the models put more weight on the reliable features, but we know of no investigation into this
%         \end{itemize}
%     \item we think the topic of non-public classification models has not received sufficient attention so far
%     \item there is also clearly a shortage of good benchmarks and realistic datasets
% \end{itemize}

We have given an overview of the field of performative prediction.
We split the problem into two major parts: knowing the distribution map and optimizing the model.
For the distribution map, we described the mechanisms that are used in the literature , characterized the categories that distribution maps can fall into and discussed why collecting datasets is difficult.
The model optimization approaches were categorized by the level of access to the distribution map that they require and by whether they target the performatively optimal or stable point.

In Section~\ref{sec:related-research-areas}, we listed several avenues for the transfer of ideas between fields.
We think there are likely many more setups that fall under the performative prediction setup, because it is a very broadly-defined one.
It is in fact so broad that it encompasses scenarios that likely should not be solved with the method discussed here.
\citet{miller2021outside} discuss the example of an election forecast (Example 3.2 in their paper):
if voters change how they vote in response to an election forecast, then that is performative prediction.
However, we would not recommend anyone try to resolve this problem by building a model of how voters react to forecasts (thereby defining a distribution map) and then using, e.g., RRM to make the election forecast take into account the voter behaviour.
While this would perhaps result in a more accurate forecast, an election forecast should preferably be kept simple --- so that it can be audited --- and free from any attempts to manipulate voters.

Often, it is also the case that while a scenario falls under performative prediction and \emph{could} be solved with performative prediction methods, the concrete details of the scenario allow a different, more practical solution.
Consider, for example, the case of the traffic prediction:
The problem statement here was that if a route is predicted to be the fastest and many people therefore take it, it might become slow instead --- for example, if the predicted fastest route is a small cross-country road.
We could solve this by building a model of driver behaviour --- thereby defining a distribution map --- and trying to anticipate their reaction to the prediction in order to make a more accurate prediction.
However, traffic pattern prediction has the advantage that it has a wealth of historical data;
it is likely that a model can be trained on historical data such that it makes accurate, non-naïve predictions about actual travel times based on the road properties of the alternative route and the nature of the blockage on the main road.
Such a sophisticated model may need to address performative prediction internally, but as the model deployer we do not have to worry about it.
Another reason for why we have more options in this case is that the traffic prediction and the drivers are not adversarial.
This means for example that the car navigation system can offer the user multiple route options and let the user randomly choose one, without worrying that the user chooses the ``worst'' option somehow.

These issue make us think that the scope of the field should be more precisely defined.
Our question is:
Is it possible to define the performative prediction setup such that it covers almost only those scenarios where the methods discussed in this paper are the most appropriate choice?

Another avenue of future research is examining exactly how methods that address performative prediction achieve the desired performatively stable or optimal result.
Do they put less weight on unreliable features?
Are they able to internally reconstruct the base features (in a setting belonging to the base-distribution family)?
These are pertinent questions for judging the robustness of these methods.

Another neglected area seems to us the situation where the users only have an imprecise idea of what the deployed classifier is.
This case was discussed in Section~\ref{sssec:dependence-on-the-model-weights}
and we consider it a quite realistic setup,
but the existing works in this area do not provide a conclusive solution to it.

Finally, the field suffers from a clear shortage of standardized benchmarks and more realistic datasets.
Due to the access pyramid (Fig.~\ref{fig:access-pyramid}),
not all methods can be applied to the same dataset --- e.g., if the dataset does not provide a full mathematical model of the distribution map, the methods in the top of the pyramid cannot be applied --- but the situation would already much improved if there was a standard dataset for each level of the pyramid.

%%% Local Variables:
%%% mode: LaTeX
%%% TeX-master: "../main"
%%% End:
    % Section 7: Discussion
%%
%% The acknowledgments section is defined using the "acks" environment
%% (and NOT an unnumbered section). This ensures the proper
%% identification of the section in the article metadata, and the
%% consistent spelling of the heading.
\section*{Acknowledgements}
We thank Myles Bartlett for the valuable discussions on the subject.

This research was funded by the European Union.  Views and opinions expressed are however those of the author(s) only and do not necessarily reflect those of the European Union or the European Health and Digital Executive Agency (HaDEA). Neither the European Union nor the granting authority can be held responsible for them.
% BayesianGDPR
J.S.B., and N.Q. have been supported in part by the European Research Council under the European Union’s Horizon 2020 research and innovation programme Grant Agreement no. 851538 - BayesianGDPR. 
% TANGO
T.K. and N.Q. have been supported in part by the Horizon Europe research and innovation programme Grant Agreement no. 101120763 - TANGO.
% BCAM
J.S.B., J.A.L. and N.Q. are also supported by BCAM Severo Ochoa accreditation CEX2021-001142-S/MICIN/AEI/10.13039/501100011033.

%%
%% The next two lines define the bibliography style to be used, and
%% the bibliography file.
\bibliography{references}

%%
%% If your work has an appendix, this is the place to put it.
\clearpage
\appendix
\section{Definitions}%
\label{appendix:definitions}
\begin{definition}[Joint smoothness]
    We say that the loss \(\ell(\theta;z)\) is \(\beta\)-\emph{jointly smooth}
    if the gradient is \(\beta\)-Lipschitz in \(\theta\) and in \(z\).
    That is:
    \begin{align}
        \|\nabla\ell(\theta; z)-\nabla\ell(\theta'; z)\|&\leq \beta\|\theta-\theta'\|\\
        \|\nabla\ell(\theta; z)-\nabla\ell(\theta; z')\|&\leq \beta\|z-z'\|
        \label{eq:jointly-smooth}
    \end{align}
    for all \(\theta, \theta'\in\Theta\) and \(z, z'\in Z\) with \(Z:=\bigcup_{\theta\in\Theta}\operatorname{supp}(\mathcal{D}(\theta))\).
\end{definition}

\section{Stackelberg games}\label{appendix:stackelberg-games}
Performative prediction is very closely related to the concept of
\emph{Stackelberg games} \cite{stackelberg2010market,julien2018stackelberg,paruchuri2008playing} originating in economics.
The following is a brief description of Stackelberg games
and how it can be re-framed as a special case of performative prediction.

A Stackelberg game involves two kinds of players (e.g., companies):
a leader and one or more followers.
Both kinds of players can take action to influence a quantity that is of interest for everyone.
In economics, this quantity is usually the unit price of a product
that both the leader and the follower are producing.
But when applying the concept to predictions,
the quantity will be something like the rate of true negatives in the predictions.
In that case,
the leader -- the organization making the predictions -- wants a \emph{high} rate of true negatives,
while the followers -- the users who are affected by the predictions -- would prefer a \emph{low} rate of true negatives (and therefore a high rate of false positives).

If we say that \(a_\ell\) is the action of the leader and \(a_0, \dots, a_n\) are the actions of the followers,
and \(P\) is the quantity of interest (e.g., a unit price or a performance metric),
then we can express the situation as follows:
\begin{equation}
  P(a_\ell, a_0, \dots, a_n)
\end{equation}
That is, the final result depends on the actions of all the players.

In the economics context, the action usually corresponds to deciding the
\emph{quantity} that the firm wants to produce of the product.
And $P$ then describes the \emph{market price} -- based on supply and demand --
that results from the total quantity that all the firms are producing in aggregate.

In our situation, the action of the leader would be the deployment of a specific classifier,
and the actions of the followers would be the strategic manipulation of their features.

The defining characteristic of Stackelberg games is that the leader moves
\emph{first};
e.g., it decides the quantity of its products first, which gives it an advantage.
Furthermore, the leader knows that the followers will act immediately after the leader has acted,
and it is assumed that the leader knows on what basis the followers will make their decision.
In economics, it is usually assumed that the followers want to maximize profit
and so the followers will choose the quantity of their products such
that they maximize profit \emph{given} the information about the quantity that the leader is producing.
This, in turn, will affect the profit of the leader.

The leader therefore is incentivized to model how the followers will respond,
and to take this into account when taking the first action.

In economics terms, each player has their own profit function which they want to maximize: 
$\Pi_\ell$ for the leader, and $\Pi_i$ for the followers.
In a simple model, the followers have a profit function of the form:
\begin{equation}
  \Pi_i(a_\ell, a_0, \dots, a_n) := P(a_\ell + a_0 + \cdots + a_n) \cdot a_i - C_i(a_i),
\end{equation}
where \(P(\dots)\cdot a_i\) could, for instance, denote a company's revenue
(unit market price times quantity of product)
and \(C_i(a_i)\) is the cost of producing that many products.
In the prediction situation, the cost can be the cost of changing the features.

Now, because the leader moves first, \(a_\ell\) is fixed by the time the followers have to decide,
so \(a_\ell\) can be assumed as given when maximizing $\Pi_i$.
However, $\Pi_i$ also depends on the other followers choice,
which makes the maximization problem very complicated (essentially,
each follower will try to predict what the other followers will do).

So, for simplicity, we will assume now that all the followers act as one, and
that there are effectively only two players: the leader and the follower.
The latter with action $a_0$.
The optimal choice for $a_0$ depends, as mentioned, on the action of the leader, \(a_\ell\):
\begin{equation}
  a_0^*(a_\ell) = \argmax_{a_0} \Pi_0(a_\ell, a_0)
\end{equation}
The leader takes this into account and optimizes the following:
\begin{equation}
  a_\ell^* = \argmax_{a_\ell} \Pi_\ell(a_\ell, a_0^*(a_\ell))
\end{equation}
That is: assuming the follower reacts in the optimal way,
what action maximizes the profit of the leader, \(\Pi_\ell\)?

It can be shown that \(a_\ell^*\) and \(a_0^*(a_\ell^*)\) is the Nash equilibrium,
meaning no player has a reason to deviate from this if no one else changes their action.

Translating this into the performative prediction setup,
the action of the leader, \(a_\ell\), corresponds to the model weights, \(\theta\);
the optimal action of the follower, \(a_0^*(a_\ell)\), is the distribution map, \(\gD(\theta)\); 
the profit of the leader, \(\Pi_\ell\), is akin to the decoupled performative risk, \(\mathrm{Risk}(\theta, Q)\), with the difference that the second argument is the model that induces the distribution instead of the distribution itself.
The Nash equilibrium corresponds to performative optimality.

% TODO
% \begin{itemize}
%     \item mention the difference that in Stackelberg games, the players are players on the same level, but in performative prediction, the ``leader'' is usually an institution that has more power than the ``followers''
%     \item mention that economists have focused on analytic solutions which we can't use in machine learning
% \end{itemize}

\section{Descriptions of commonly-used datasets in performative prediction}%
\label{appendix:commonly-used-datasets}
\subsection{GiveMeSomeCredit with logistic regression}%
\label{sssec:givemesomecredit-with-logistic-regression}
The most commonly-used existing dataset in the literature is GiveMeSomeCredit~\cite{GiveMeSomeCredit}
with logistic regression as the classifier.
This dataset contains features describing the income and the credit history of 250,000 individuals
--- e.g., the number of times they have been at least 90 days late in their payments ---
and the goal is to predict whether someone will default within 2 years.

For the base distribution, one can simply choose a uniform distribution over the samples in the dataset,
or something more sophisticated, like a class-balanced distribution.
As the labels in the dataset correspond to the default risk,
the desirable outcome is to get a negative label,
so the utility function is chosen to be the negative of the output of the logistic regression model:
\(u(x, \theta) = -\theta^T x\), where \(\theta\) is the weight vector.

The final required element is the cost function, which is defined to be the squared distance of the feature vectors:
\begin{equation}
  c(x, x')=\tfrac1{2\epsilon}{\|x-x'\|}^2_2~.
  \label{eq:quadratic-cost-function}
\end{equation}
The parameter \(\epsilon\) controls how strongly to weight the cost in comparison to the utility.
In this simple case, we weight all features in the same way.
A more complex cost function could take into account that some feature values are easier to change than others
(e.g., you cannot really change the number of times you have been late in your payments in the past,
but you \emph{can} change your income).

Solving~\eqref{eq:strategic-classification} for these utility and cost functions,
we get the following expression for samples drawn from distribution \(\mathcal{D}(\theta)\):
\begin{equation}
  (x_ \mathrm{base} - \epsilon \theta, y_ \mathrm{base})\quad \text{where }(x_ \mathrm{base}, y_ \mathrm{base}) \sim \mathcal{D}_ \mathrm{base}
  \label{eq:logistic-regression-dist-map}
\end{equation}

\subsection{UCI Credit}%
\label{sssec:uci-credit}
The UCI Credit dataset~\cite{default_of_credit_card_clients_350} has 30,000 samples with a label for whether a person defaulted on a credit card payment.
There are 23 features, including education, marital status and several features about the history of past payments.
For the purposes of experiments, it is assumed that only education level and payment patterns are changeable by the user~\cite{lin2024plugin}.

\subsection{Various synthetic datasets}%
\label{sssec:various-synthetic}

\citet{izzo2021learn} define a series of distribution maps based on the normal distribution. These distribution maps include, among others, non-linear dependencies, $\mathcal{D}_\mathrm{nl}(\theta)=\mathcal{N}(\sqrt{a_1\theta+a_0};\sigma^2)$, $a_1, a_0 \in \mathbb{R}$, or mixture of Gaussians, $\mathcal{D}_\mathrm{mix}(\theta)=\gamma \mathcal{N}(a_{0,1}\theta + a_{0,0};\sigma_0^2) + (1-\gamma) \mathcal{N}(a_{1,1}\theta + a_{1,0};\sigma_1^2)$, $a_{i,i}\in \mathbb{R}$, $\sigma_i\in \mathbb{R}_{>0}$, for $i=0,1$. 
The distribution may also be multivariate.

\section{From delayed impact to performative prediction}%
\label{appendix:delayed-impact}
% https://arxiv.org/abs/2310.08349
% Novi's take: it's about fairness metrics; new fairness metric taking into account long-term effects, feedback loops; example: parental leave
The origins of the delayed impact framework are that \citet{d2020fairness} examined the long-term impact of interventions intended to improve the standing of a disadvantage group (i.e., through positive discrimination).
\citet{liu2018delayed} concluded that such interventions may have the opposite of the intended effect by leaving those groups worse off whose situation was supposed to be improved.
In particular, the impact might seem positive at first, but the \emph{delayed} impact can still be negative.
The reason is that the decisions that the institution made, have effects on the world besides the immediate decision.

The authors develop a mathematical model for the long-run effect such interventions can have.
As we will see, this mathematical model can be reformulated in terms of a distribution map.
We can then identify the delayed impact of an intervention with the \emph{equilibrium state} from performative prediction.

The paper considers a simple scenario.
The decision is based on two features:
the group membership, $j\in\{A, B\}$, and a categorical feature, $x$, which in the running example of the paper is the \emph{credit score} of a person.
A predictive model, $\tau_j(x)$, outputs the probability that a candidate with group membership $j$ and credit score $x$ should be accepted.
The institution has a utility function, $u(x)$, which captures how much the institution gains by accepting candidate with feature $x$.
In the credit example, this is the profit of the bank and can be expressed as $u(x)=u_+\rho(x)+u_-\big(1-\rho(x)\big)$
where $\rho(\cdot)$ is the probability that the loan will be repaid
and $u_+$ and $u_-$ are, respectively, the (possibly negative) profit the bank makes when the loan is repaid or defaulted on.%
A curiosity here is that this utility function, $u(x)$ does not depend on the group $j$,
which seems to imply that $j$ is not needed to determine the utility (i.e., profit).
This is counter to the usual algorithmic fairness setup where the group membership $j$ \emph{is} predictive of the prediction target, even in the presence of all other features $x$, but the problem is that we do not want to use $j$ for ethical or legal reasons.%
If we define $z=(j, x)$, then we can define the loss for the bank as the negative expected utility for sample $z$:
\begin{align}
    \ell(z; \tau) = \ell\big((j, x), \tau\big) = -\tau_j(x)\cdot u(x)~.
\end{align}
Let $\mathcal{D}_0$ be the initial data distribution over $z=(j, x)$.
The initial expected loss is
\begin{align}
    \mathcal{L}(\tau, \mathcal{D}_0)=\mathbb{E}_{z\sim\mathcal{D}_0}\big[\ell(z;\tau)\big]
\end{align}
dependent on the classifier $\tau$ and the distribution $\mathcal{D}_0$.

So far, we have only described the immediate utility, so let us consider the delayed effect.
In the credit example, the effect we are considering is how the credit score changes in response to repaid or defaulted loans.
We define $\Delta(x)$ to be the change in feature $x$ (i.e., the change in credit score) due to long-term effects, that happens after the candidate was accepted.
In the credit example, a simple model is $\Delta(x)=c_+\rho(x)+c_-\big(1-\rho(x)\big)$
where $c_+$ is the increase in credit score from a repaid loan and $c_-$ the decrease from a defaulted loan, and $\rho$ is defined as before (probability of repaid loan).
From this, we can define a distribution map of the base-distribution family, where $\mathcal{D}_0$ is the base
and $\tau_j(x)\cdot \Delta(x)$ gives us the expected change in credit score,
which we formalize with the transformation function,
$\varphi\big((j, x), \tau\big) = \big(j, x+\tau_j(x)\cdot \Delta(x)\big)$.
The distribution thus depends on \(\tau\) and the problem becomes performative.

This gives us the distribution map, but there is still a step missing in order for the setting to be considered performative prediction:
targeting the performative risk, or performative utility (the inverse of the risk).
The performative utility here corresponds to the profit at the shifted distribution.
However, in the original paper, the bank only considers the profit of the original distribution --- the bank does not look ahead.
Indeed, the point of the original paper was that contrary to expectation, short-term profit-maximizing behaviour by an institution can be beneficial for the users in the long term, where user well-being is, e.g., measured by their credit score.
It stands to reason, however, that long-term profit-maximizing behaviour by the institution would be even better for users --- such a behaviour would take into account the expected shifting of the distribution.
It is perhaps unlikely that a bank would \emph{only} consider future profit (and ignore immediate profit), but the bank could include future profit in its calculation, multiplied by a discount rate~\cite{gollier2002discounting}.
With this, the conditions for performative prediction would be satisfied:
the classifier causes a shift in the distribution (in the credit scores) and the deployer cares at least partially about the risk (or utility) at the equilibrium point.

\end{document}